\documentclass{article}

\usepackage{microtype}
\usepackage{graphicx}
\usepackage{subcaption}
\usepackage{booktabs} 
\usepackage{amssymb}
\usepackage{amsmath}
\usepackage{multirow}
\usepackage{adjustbox}
\usepackage{bbding}
\usepackage{float}
\usepackage{titletoc}
\usepackage{hyperref}
\usepackage{algorithmic}
\usepackage[most]{tcolorbox}

\usepackage{xcolor}
\usepackage{listings} 
\definecolor{mainblue}{RGB}{0, 114, 178}
\definecolor{backblue}{RGB}{240, 247, 255}

\lstdefinelanguage{json}{
    basicstyle=\small\ttfamily,
    numbers=none,
    numberstyle=\scriptsize,
    stepnumber=1,
    numbersep=8pt,
    showstringspaces=false,
    breaklines=true,
    frame=none,
    backgroundcolor=\color{backblue}, 
    string=[s]{"}{"},
    comment=[l]{:\ "},
    morecomment=[s]{"}{"},
    stringstyle=\color{blue},
    commentstyle=\color{black},
    literate=
     *{0}{{{\color{black}0}}}{1}
      {1}{{{\color{black}1}}}{1}
      {2}{{{\color{black}2}}}{1}
      {3}{{{\color{black}3}}}{1}
      {4}{{{\color{black}4}}}{1}
      {5}{{{\color{black}5}}}{1}
      {6}{{{\color{black}6}}}{1}
      {7}{{{\color{black}7}}}{1}
      {8}{{{\color{black}8}}}{1}
      {9}{{{\color{black}9}}}{1}
      {:}{{{\color{blue}:}}}{1}
      {,}{{{\color{blue},}}}{1}
      {\{}{{{\color{blue}\{}}}{1}
      {\}}{{{\color{blue}\}}}}{1}
      {[}{{{\color{blue}[}}}{1}
      {]}{{{\color{blue}]}}}{1},
}

\tcbset{
    unified_style/.style={
        enhanced, arc=2mm, boxrule=1pt, colframe=mainblue, colback=backblue,
        coltitle=white, colbacktitle=mainblue, fonttitle=\bfseries\sffamily,
        fontupper=\small\sffamily, breakable, drop shadow={black!10!white},
        attach boxed title to top left={xshift=5mm, yshift=-3mm},
        left=4mm, right=4mm, top=3mm, bottom=2mm
    }
}

\newtcolorbox{protocolbox}[1]{unified_style, title=#1}

\newcommand{\TO}{\textbf{to }}                      
\newcommand{\RETURN}{\STATE \textbf{return }}
\newcommand{\NOT}{\textbf{not }}      

\usepackage[preprint]{icml2026}

\usepackage{amsmath}
\usepackage{amssymb}
\usepackage{mathtools}
\usepackage{amsthm}

\usepackage[capitalize,noabbrev]{cleveref}

\theoremstyle{plain}

\theoremstyle{definition}

\theoremstyle{remark}

\usepackage[textsize=tiny]{todonotes}

\begin{document}

\twocolumn[
  \icmltitle{MedSAM-Agent: Empowering Interactive Medical Image Segmentation with Multi-turn Agentic Reinforcement Learning}

  \icmlsetsymbol{corresponding}{\textdagger}

  \begin{icmlauthorlist}
    \icmlauthor{Shengyuan Liu}{cuhk}
    \icmlauthor{Liuxin Bao}{cuhk}
    \icmlauthor{Qi Yang}{tencent,cas}
    \icmlauthor{Wanting Geng}{tencent,dlut}
    \icmlauthor{Boyun Zheng}{cuhk} \\
    \icmlauthor{Chenxin Li}{cuhk}
    \icmlauthor{Wenting Chen}{stanford}
    \icmlauthor{Houwen Peng}{tencent,corresponding}
    \icmlauthor{Yixuan Yuan}{cuhk,corresponding}
  \end{icmlauthorlist}

  \icmlaffiliation{cuhk}{Chinese University of Hong Kong, Hong Kong SAR, China}
  \icmlaffiliation{tencent}{Hunyuan Group, Tencent}
  \icmlaffiliation{cas}{Institute of Automation, the Chinese Academy of Sciences, Beijing, China}
  \icmlaffiliation{dlut}{Dalian University of Technology, Dalian, China}
  \icmlaffiliation{stanford}{Stanford University, Stanford, USA}

  \icmlcorrespondingauthor{Yixuan Yuan}{yxyuan@ee.cuhk.edu.hk} 
  \icmlcorrespondingauthor{Houwen Peng}{henryllpeng@tencent.com} 

  \vskip 0.3in
]


\printAffiliationsAndNotice{}  

\begin{abstract}
Medical image segmentation is evolving from task-specific models toward generalizable frameworks. Recent research leverages Multi-modal Large Language Models (MLLMs) as autonomous agents, employing reinforcement learning with verifiable reward (RLVR) to orchestrate specialized tools like the Segment Anything Model (SAM). However, these approaches often rely on single-turn, rigid interaction strategies and lack process-level supervision during training, which hinders their ability to fully exploit the dynamic potential of interactive tools and leads to redundant actions. To bridge this gap, we propose MedSAM-Agent, a framework that reformulates interactive segmentation as a multi-step autonomous decision-making process. First, we introduce a hybrid prompting strategy for expert-curated trajectory generation, enabling the model to internalize human-like decision heuristics and adaptive refinement strategies. Furthermore, we develop a two-stage training pipeline that integrates multi-turn, end-to-end outcome verification with a clinical-fidelity process reward design to promote interaction parsimony and decision efficiency. Extensive experiments across 6 medical modalities and 21 datasets demonstrate that MedSAM-Agent achieves state-of-the-art performance, effectively unifying autonomous medical reasoning with robust, iterative optimization. Code is available \href{https://github.com/CUHK-AIM-Group/MedSAM-Agent}{here}.

\end{abstract}

\begin{figure}[t]
  \centering 
  \includegraphics[width=\linewidth]{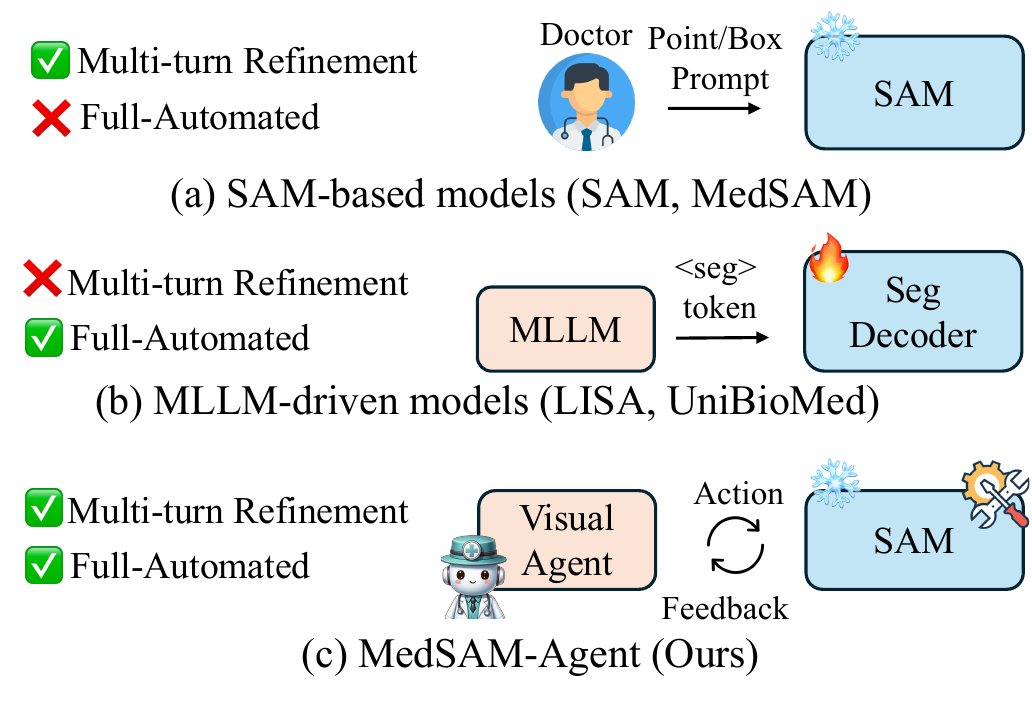}
   \caption{Comparison of medical image segmentation paradigms. (a) SAM-based models (e.g., SAM, MedSAM) require continuous manual prompting via points or bounding boxes. (b) MLLM-driven models (e.g., LISA, UniBioMed) employs MLLM with specialized seg decoders and \texttt{<seg>} tokens. (c) Ours MedSAM-Agent functions as an autonomous visual agent that performs multi-turn refinement through iterative feedback and tool interaction, emulating the professional decision-making process.}
   \label{fig:fig1}
\end{figure} 

\section{Introduction}
\label{sec:intro}
Medical image segmentation stands as a foundational task in clinical computer vision, underpinning critical applications such as early disease diagnosis, surgical planning, and treatment response assessment. Traditional AI-driven segmentation approaches, including UNet \cite{unet} and its variants \cite{transunet,nnunet,ukan,unet++}, have demonstrated impressive performance in medical segmentation scenarios. However, these methods are primarily tailored to specific tasks or imaging modalities, posing significant challenges when generalizing to new tasks that were not encountered during model training. The emergence of the Segment Anything Model (SAM) \cite{SAM, SAM2} represented a pivotal breakthrough, enabling high-quality segmentation through interactive human prompts (e.g., points, bounding boxes, or masks) \cite{MedSAM,I-MedSAM,EnhancingSAMMed,medsam2,MedSAMAdapter,3dsam}. Nevertheless, these SAM-derived interactive models remain inherently dependent on expert prompting (Fig.~\ref{fig:fig1}a), which prevents them from achieving autonomous, generalized segmentation without manual intervention. 

Meanwhile, Multi-modal Large Language Models (MLLMs) \cite{medgemma,Lingshu,hulumed,Huatuo-Vision} have shown remarkable perception and reasoning abilities in the medical domain, especially in visual question answering  \cite{omnimedvqa,gmaimmbench,slake,VQA-RAD,pathvqa,endobench} and report generation \cite{braingpt,llavarad,pathbench,chen_reportgen,dermogpt} tasks. While early attempts \cite{UniBiomed,MedPLIB,citrusV} integrate MLLMs for segmentation via implicit tokens and additional pixel decoders (Fig.~\ref{fig:fig1}b), these methods often compromise semantic generalization by altering the original output space and shifting away from language-based outputs. Recent advancements \cite{mmedagent,nath2025vila,aura} have shifted toward employing MLLMs as agents for invoking SAM tools, utilizing Reinforcement Learning from Verifiable Rewards (RLVR) during the post-training stage to integrate the high-level reasoning of MLLMs with the robust interactive segmentation of SAM by internalizing the autonomous tool-using.

However, existing methods still suffer from two major limitations: First, interaction strategies remain simplistic and inefficient. Current studies either rely on single-turn prompts \cite{Seg-R1,Seg-Zero,SAM-R1,mmedagent}, treating SAM as a static segmentor rather than an iterative agent, or adhere to rigid, point-only trajectories \cite{SegAgent,ibisagent}. These point-centric paradigms lack the spatial flexibility to adaptively encompass morphological heterogeneity. In voluminous or ambiguously bounded cases, such methods fail to leverage bounding boxes as a human-like anchor for global context, leading to suboptimal step-wise refinement and under-utilized iterative potential. Second, the reinforcement learning process lacks process-level supervision. Existing RLVR frameworks \cite{Seg-R1,SAM-R1,Seg-Zero,Ophiuchus,VisionReasoner,pixelreasoner} focus predominantly on terminal outcomes, such as the final segmentation accuracy (e.g., Dice or IoU), while neglecting the efficiency and logical coherence of intermediate actions. Without specific rewards to incentivize action parsimony and per-step effectiveness, the agent may take unnecessary steps that do not contribute to mask refinement, which ultimately compromises both training convergence efficiency and final inference performance.

To address these limitations, we propose MedSAM-Agent, a framework that enables MLLMs to emulate human annotators by treating the use of interactive segmentation tools as a multi-step decision-making process (Fig.~\ref{fig:fig1}c). First, we implement a hybrid prompting strategy to generate diverse, high-quality expert trajectories, enabling the model to internalize human-like decision heuristics and adaptive refinement strategies by boxes and points. Then we develop a two-stage training pipeline that integrates multi-turn, end-to-end outcome verification with a clinical-fidelity process reward design to promote interaction parsimony and decision efficiency. This framework jointly optimizes for segmentation accuracy and procedural efficiency, ensuring the agent learns to make precise, non-redundant decisions. Consequently, MedSAM-Agent achieves state-of-the-art performance across 6 medical modalities and 21 datasets, demonstrating exceptional cross-modal generalization and tool-use versatility. Our key contributions are summarized as follows:

\begin{itemize}
    \item We introduce the MedSAM-Agent framework, which reformulates medical image segmentation by transitioning from static pixel-wise classification to a dynamic decision-making paradigm.
    \item We develop a hybrid prompting strategy for expert-curated trajectory generation, followed by a two-stage training pipeline optimized via a clinical-fidelity process reward design.
    \item We conduct comprehensive experiments across 6 medical modalities that demonstrate superior segmentation performance and robust tool-agnostic generalization.
\end{itemize}

\section{Related Works}
\label{sec:related}

\subsection{Medical Image Segmentation}
Traditional medical image segmentation methods like UNet \cite{unet}, nn-UNet \cite{nnunet}, and other representative architectures \cite{ukan, unet++, transunet, pranet} are inherently task-specific, which require dedicated training for each individual segmentation task, thus exhibiting limited generalization across diverse medical imaging scenarios \cite{universeg,UniSeg}. The advent of the SAM \cite{SAM, SAM2} has sparked renewed hope for achieving generalizable medical image segmentation. Consequently, numerous SAM-derived models \cite{MedSAM,I-MedSAM,EnhancingSAMMed,MedSegX,medsam2,MedSAMAdapter,sam-med3d,biomedparse} have been fine-tuned on large-scale medical image datasets. However, these interactive segmentation models typically rely on explicit human-provided prompts (e.g., points, boxes, or masks) to guide the segmentation process, limiting their autonomy. Parallel to this, the advancement of MLLMs has enabled text-guided medical image segmentation \cite{MedPLIB,MedRegA,unimedvl}, such as Citrus-V \cite{citrusV} and UniBiomed \cite{UniBiomed}. Despite their flexibility in leveraging textual instructions, these MLLM-driven methods are constrained by the inherent limitations of MLLMs in capturing fine-grained pixel-level details, and they lack the iterative refinement capability that characterizes SAM-based interactive segmentation frameworks. In contrast, our proposed method addresses these gaps by mimicking the annotation trajectories of human experts through the integration of interactive segmentation tools, enabling both autonomous guidance and iterative optimization.

\subsection{Reinforcement Learning in MLLM}

Driven by the success of DeepSeek-R1 \cite{GRPO}, reinforcement learning (RL) has fundamentally reshaped the training paradigms for large language models \cite{ppo,DAPO,GSPO,REINFORCE}. Recent studies \cite{Lmm-r1, vlm-r1, video-r1, VisualRFT, Seg-Zero,VisualRFT} have demonstrated that simple yet verifiable rewards can effectively extend a model's reasoning capabilities from text-only tasks to multi-modal scenarios, a transition that has also proven effective in specialized medical tasks \cite{reasoning_survey, med-r1, medvlm-r1, Med-rlvr, medthinkiter, medground-r1}. For fine-grained understanding tasks, milestone developments like OpenAI-o3 \cite{openai-o3} have introduced the concept of ``thinking with images", leveraging iterative zoom-in mechanisms \cite{DeepEyes, minio3, shen2025zoomeye, pixelreasoner, thyme} or external tool invocation \cite{OpenThinkIMG, SAM-R1, Seg-R1, vtools-r1, thinkingwithvideos} to actively refine visual perception. In the medical domain, while similar efforts \cite{medseg-r,medreasoner,mmedagent,medagent-pro} have attempted to integrate tool-use, existing medical MLLMs \cite{mmedagent,ibisagent,Ophiuchus} predominantly operate within a single-turn interaction paradigm or rely on monotonic clicking strategies that lack strategic diversity. Consequently, their potential to function as autonomous agents capable of dynamic, multi-step action planning remains underdeveloped.

\section{Methods}

\begin{figure*}[htb]
  \centering \includegraphics[width=\linewidth]{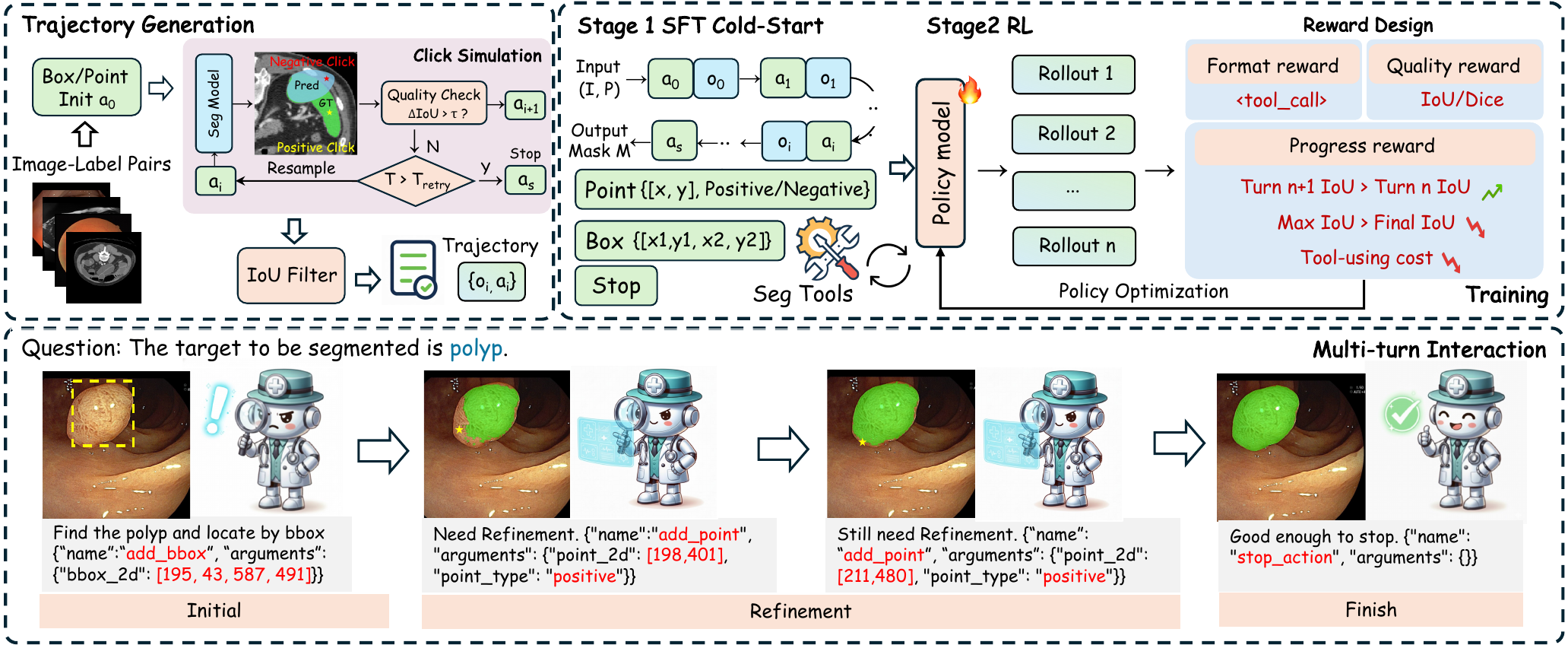}
   \caption{\textbf{Overview of MedSAM-Agent.} We develop a hybrid prompting strategy for expert-curated trajectory generation that transforms image-label pairs into high-quality interaction sequences via simulated clicks and IoU-based filtering. Then these trajectories support a two-stage training pipeline, stage-1 SFT cold-start for initial capability and stage-2 RL optimized by a fine-grained reward design. MedSAM-Agent can autonomously select between box and point tools and execute the ``stop" action once the refinement is complete.}
   \label{fig:main}
\end{figure*} 

\subsection{Expert-Curated Trajectory Generation}

Given an image $I$ and a segmentation target prompt $P$, medical experts commonly employ interactive segmentation tools: based on the original image $I$ and the real-time updated mask $M$, they iteratively provide positive click points (marking regions confirmed as the target) and negative click points (excluding non-target regions). This iterative interaction facilitates the efficient generation of high-quality masks that meet clinical annotation standards. We formalize this procedure as a multi-step decision-making process. Given the prompt $P$ and input image $I$, the policy model iteratively generates an action $a_t$. This action interacts with the environment by invoking the image segmentation tool, resulting in a new observation $o_t$. This observation is appended to the interaction history and fed back to the policy model $\pi_\theta(a_t|s_t, I, P)$. The components are detailed as follows:

\textbf{Action $a_t$:} To simulate expert behavior, we define the action space to encompass bounding box operations, point-wise clicks, and a stop signal. The box operation is represented as $a_t = [x_1, y_1, x_2, y_2]$ to provide a global spatial prior, while a click operation consists of a coordinate $a_t = [x_1, y_1]$ paired with an attribute $\alpha \in \{+1, -1\}$ to denote positive or negative refinement; finally, the stop operation is triggered to terminate the sequence once the segmentation reaches the desired fidelity.

\textbf{Observation $o_t$:} In agentic RL, the observation $o_t$ is obtained based on the parameters indicated by the corresponding action $a_t$. Concretely, it is the updated mask $M_t$ predicted by the interactive segmentation network $F_{seg}$. Prior to the execution of a stop action, the interactive model maintains a record of all previous actions, with the updated mask computed as $M_{t} = F_{seg}(I, o_{t-1}, a_t)$, where $M_t$ directly serves as $o_t$. These observation tokens are appended to the ongoing rollout sequence and fed back into the model as input for the subsequent step. When the action $a_t = stop\_action$, no tool invocation is performed; instead, $o_t$ directly inherits the result from the previous iteration $o_{t-1}$ and is output as the final predicted mask.

\textbf{State $s_t$:}  At each step $t$, the state $s_t$ is defined as:
\begin{equation}
s_t = \{(a_0, o_0), (a_1, o_1), \ldots, (a_{t-1}, o_{t-1})\}
\end{equation}
Given the state $s_t$, the action $a_t \sim \pi_\theta(a \mid s_t)$ is sampled from the MLLM policy $\pi_\theta$, serving as the next input token. This long sequence continues to interleave until either the stop action is generated or the maximum number of tool calls is reached. The objective of our framework is to develop an MLLM capable of executing a segmentation policy $\pi_\theta(a_t|s_t, I, P)$ that emulates the decision-making process of medical experts. To facilitate this, we construct a trajectory dataset $D_{\text{traj}} = \{(I, M_{\text{target}}, P, s_t)\}$, derived from an initial dataset $D_{\text{seg}} = \{(I, M_{\text{target}}, P)\}$.

\textbf{Hybrid Prompting Strategy.} While existing methods \cite{SegAgent,ibisagent} rely on a rigid point-wise simulation driven by pixel-level discrepancies, they typically restrict the action space to sequential clicks. Such a limitation fails to capture the multi-modal nature of human workflows, where practitioners often initiate segmentation by defining an ROI. To bridge this gap, we propose a hybrid prompting strategy (Fig.~\ref{fig:main}) that offers greater spatial flexibility. This strategy incorporates both Box-to-Point and Sequential-Click paradigms to better mimic clinical expertise. In the Box-to-Point workflow, the trajectory is initialized with a bounding box action generated by extracting the axis-aligned rectangle of $M_{\text{target}}$ with a controllable random jitter to simulate human imprecision. For subsequent refinement, the agent identifies the most significant False Negative (FN) and False Positive (FP) components. By applying a distance transform to these error regions, corrective clicks are sampled at the centroids of the largest error clusters, ensuring each action addresses the most salient morphological defects.

To guarantee the quality and efficiency of the synthesized trajectories, we implement a progress-constrained sampling mechanism. We contend that an expert-level trajectory must consist of informative and effective actions; consequently, we enforce a constraint where each simulated action must yield an incremental IoU gain ($\Delta \text{IoU}$) exceeding a predefined threshold $\tau$. To prevent the inclusion of sub-optimal or redundant interactions, a retry mechanism is introduced: if a candidate action fails to satisfy the progress threshold, the simulator performs iterative resampling up to $N$ trials to identify a more constructive interaction.Following the generation of all sequences, a global IoU filter is applied to both paradigms to prune trajectories that fail to reach a target performance threshold. This validation process effectively filters out stochastic noise and ensures that $D_{\text{traj}}$ is composed of high-quality, monotonically improving sequences, thereby providing a robust supervisory signal for policy optimization. By integrating these components, we successfully construct the expert-curated trajectories $D_{\text{traj}}$. The details of the trajectory generation algorithm are shown in the Appendix.

\subsection{Two-stage Training Pipeline}
Building upon these trajectories, we employ a two-stage training pipeline: Supervised Fine-Tuning (SFT) serves as the initial cold-start, followed by Reinforcement Learning with Verifiable Rewards (RLVR) in the second stage to further refine the agent's decision-making policy. The overall training pipeline is shown in Fig.~\ref{fig:main}.
\subsubsection{SFT Cold-start}

To bridge the gap between natural language reasoning and tool interaction, we format the trajectory data into a structured tool-calling schema. Each expert action $a_t$ is converted into a specialized token sequence enclosed within \texttt{<tool\_call>} tags, resulting in a SFT dataset $D_{\text{SFT}}$ comprising 449K samples. The system prompt $P$ encompasses a comprehensive description of the task requirements and action space, alongside a semantic definition of the target object. This design ensures the model maintains a global objective while grounding its actions in the specific visual and textual context of the segmentation target. The details of the prompt design are shown in the Appendix.

As each expert-curated action is tokenized into a discrete sequence, the model is trained to predict the next token in an auto-regressive manner. Formally, the optimization objective is defined by the following negative log-likelihood loss:
\begin{equation}
\mathcal{L}_{\text{SFT}}(\theta) = - \mathbb{E}_{(I, P, y) \sim D_{\text{sft}}} \left[ \sum_{t=1}^{|y|} \log \pi_\theta(y_t | I, P, y_{<t}) \right]
\end{equation}
where $y$ denotes the target token sequence representing the expert trajectory $s_T$, $y_t$ is the $t$-th token of the sequence, and $y_{<t}$ represents the preceding context including the image $I$, the text prompt $P$, and previously generated tokens. To provide the model with a precise perception of the current segmentation state, the mask generated at each step is overlaid on the original image $I$ as a dedicated visual prompt, which is then re-encoded by the vision encoder to inform the subsequent action. This iterative visual feedback loop enables the agent to dynamically observe the refined boundaries and localize remaining errors throughout the trajectory.

This cold-start phase is foundational for the subsequent reinforcement learning stage; by minimizing the imitation loss, the model aligns its initial policy distribution with the decision-making heuristics of medical experts, effectively acquiring fundamental visual grounding capabilities. This strategic initialization ensures that the agent enters the reinforcement learning phase with a pre-established competency in autonomous tool invocation and spatial reasoning.

\subsubsection{Clinical-fidelity Process Reward Design}
Unlike prior approaches \cite{Seg-Zero, SAM-R1, Seg-R1} that predominantly rely on outcome-based rewards for single-pass predictions, we propose a multi-dimensional reward framework. This design meticulously balances final segmentation fidelity with the strategic efficiency of the interactive process, fostering a policy that is both accurate and resource-conscious.

\textbf{Format Reward ($R_{fmt}$):} The format reward $R_{fmt}$ is engineered to enforce adherence to the interactive protocol through a partial-credit mechanism. It rewards two fundamental behaviors: (1) the active deployment of interaction primitives (e.g., $add\_bbox$ or $add\_point$) and (2) the execution of a definitive $stop\_action$ to terminate the sequence. By assigning $0.5$ points to each criterion, we prevent the model from collapsing into infinite loops, ensuring the agent learns the logical boundaries of task completion.

\textbf{Quality Reward ($R_{qual}$):} The core objective is to maximize the final intersection between the predicted mask and the ground truth. We define the quality reward $R_{qual}$ as a weighted ensemble of the IoU and the Dice Coefficient:
\begin{equation}R_{qual} = w_{iou} \cdot \text{IoU}_{final} + w_{dice} \cdot \text{Dice}_{final}
\end{equation}
$w_{iou}$ and $w_{dice}$ are weighted hyper-parameters. While IoU provides a standard geometric consistency measure, the inclusion of the Dice coefficient is critical for medical contexts, as it offers higher sensitivity to small-scale lesions and anatomical structures, thereby mitigating the penalty variance in class-imbalanced scenarios.

\begin{table*}[t!]
\centering
\caption{\textbf{Quantitative comparison across six medical imaging modalities.} The results report the mean Dice and IoU scores for each specific modality and the overall average. \textbf{Point} and \textbf{Box} indicate the interactive prompts derived from ground-truth center points and bounding boxes, used to evaluate single-round segmentation performance. \textbf{Bold} and \underline{underlined} values represent the best and second-best performance among non-interactive methods, respectively.}
\setlength{\tabcolsep}{2.5pt}
\begin{adjustbox}{width=\textwidth,center}
\begin{tabular}{lcccccccccccccc}
\toprule
\multirow{2}{*}{\textbf{Method}} & \multicolumn{2}{c}{\textbf{CT}} & \multicolumn{2}{c}{\textbf{MRI}} & \multicolumn{2}{c}{\textbf{X-Ray}} & \multicolumn{2}{c}{\textbf{Ultrasound}} & \multicolumn{2}{c}{\textbf{Fundus}} & \multicolumn{2}{c}{\textbf{Endoscopy}} & \multicolumn{2}{c}{\textbf{Avg}}\\
\cmidrule(lr){2-3} \cmidrule(lr){4-5} \cmidrule(lr){6-7} \cmidrule(lr){8-9} \cmidrule(lr){10-11} \cmidrule(lr){12-13} \cmidrule(lr){14-15}    
& Dice & IoU & Dice & IoU & Dice & IoU & Dice & IoU & Dice & IoU & Dice & IoU & Dice & IoU \\
\midrule
\textit{SAM based methods} \\
SAM2-Point~\cite{SAM2} & 0.610 & 0.531 & 0.488 & 0.412 & 0.546 & 0.422 & 0.528 & 0.408 & 0.362 & 0.265 & 0.659 & 0.597 & 0.532 & 0.439 \\
SAM2-Box~\cite{SAM2} & 0.837 & 0.754 & 0.838 & 0.748 & 0.827 & 0.733 & 0.871 & 0.779 & 0.717 & 0.574 & 0.922 & 0.872 & 0.835 & 0.743 \\
MedSAM-Point~\cite{medsam2} & 0.747 & 0.640 & 0.667 & 0.584 & 0.843 & 0.776 & 0.591 & 0.467 & 0.414 & 0.346 & 0.822 & 0.736 & 0.681 & 0.592 \\
MedSAM-Box~\cite{medsam2} & 0.847 & 0.759 & 0.854 & 0.768 & 0.897 & 0.836 & 0.890 & 0.806 & 0.846 & 0.743 & 0.925 & 0.872 & 0.876 & 0.797 \\
IMISNet-Point~\cite{IMISNet} & 0.674 & 0.580 & 0.678 & 0.556 & 0.578 & 0.449 & 0.489 & 0.371 & 0.677 & 0.528 & 0.781 & 0.681 & 0.646 & 0.527 \\
IMISNet-Box~\cite{IMISNet} & 0.805 & 0.727 & 0.849 & 0.758 & 0.706 & 0.628 & 0.569 & 0.483 & 0.841 & 0.733 & 0.905 & 0.846 & 0.779 & 0.696 \\
\midrule
\textit{MLLM based methods} \\
LISA~\cite{LISA} & 0.088 & 0.051 & 0.110 & 0.063 & 0.322 & 0.230 & 0.385 & 0.280 & 0.036 & 0.019 & 0.282 & 0.222 & 0.204 & 0.144 \\
GLAMM~\cite{GALLM} & 0.088 & 0.053 & 0.106 & 0.061 & 0.285 & 0.194 & 0.402 & 0.294 & 0.027 & 0.014 & 0.245 & 0.179 & 0.192 & 0.132 \\
HyperSeg~\cite{hyperseg} & 0.098 & 0.069 & 0.153 & 0.101 & 0.336 & 0.231 & 0.434 & 0.336 & 0.023 & 0.012 & 0.396 & 0.342 & 0.240 & 0.182 \\
Seg-R1~\cite{Seg-R1} & 0.099 & 0.064 & 0.126 & 0.075 & 0.417 & 0.318 & 0.490 & 0.382 & 0.455 & 0.325 & 0.533 & 0.471 & 0.353 & 0.272 \\
MedPLIB~\cite{MedPLIB} & 0.052 & 0.038 & 0.177 & 0.136 & 0.115 & 0.075 & 0.088 & 0.056 & 0.298 & 0.201 & 0.132 & 0.097 & 0.144 & 0.101 \\
UniBiomed~\cite{UniBiomed} & \underline{0.724} & \underline{0.634} & \textbf{0.807} & \textbf{0.716} & \underline{0.817} & \underline{0.748} & \underline{0.736} & \underline{0.618} & 0.794 & 0.668 & 0.778 & 0.704 & \underline{0.776} & \underline{0.681} \\
Citrus-V\cite{citrusV} & 0.363 & 0.302 & 0.326 & 0.258 & 0.624 & 0.536 & 0.134 & 0.094 & 0.782 & 0.655 & 0.666 & 0.616 & 0.482 & 0.410 \\
\midrule
Qwen3-VL+MedSAM2 (RL only) & 0.401 & 0.346 & 0.647 & 0.559 & \textbf{0.834} & \textbf{0.779} & 0.718 & 0.600 & 0.775 & 0.644 & 0.695 & 0.631 & 0.678 & 0.593 \\
Ours-IMISNet & \textbf{0.732} & \textbf{0.654} & 0.783 & 0.685 & 0.798 & 0.723 & 0.658 & 0.547 & \underline{0.803} & \underline{0.678} & \underline{0.803} & \underline{0.736} & 0.763 & 0.670 \\
Ours-MedSAM2 & 0.720 & 0.633 & \underline{0.793} & \underline{0.701} & \underline{0.833} & \underline{0.775} & \textbf{0.793} & \textbf{0.685} & \textbf{0.813} & \textbf{0.692} & \textbf{0.811} & \textbf{0.744} & \textbf{0.794} & \textbf{0.705} \\
\bottomrule
\end{tabular}
\end{adjustbox}
\label{tab:tab1}
\end{table*}

To optimize the agent’s decision-making trajectory, we introduce a process reward comprised of three distinct components designed to foster intelligent interaction logic:

\textbf{Progressive Improvement Bonus ($R_{imp}$):} This component incentivizes the agent to achieve monotonic quality gains across iterations. By rewarding the cumulative positive deltas in IoU, we discourage redundant or ineffective interactions that fail to contribute to mask refinement.
\begin{equation}
R_{imp} = \sum_{t=1}^{N-1} \max(0, \text{IoU}_t - \text{IoU}_{t-1})
\end{equation}
\textbf{Overshoot Penalty ($R_{over}$):} This penalty is designed to encourage efficient termination and prevent redundant interactions. Since the model lacks access to ground-truth IoU during inference, it must internalize the ability to decide when to conclude an interaction based solely on visual cues. By penalizing any quality degradation following the peak IoU ($\text{IoU}_{max} - \text{IoU}_{final}$), during training, $R_{over}$ effectively disciplines the agent to recognize the point of diminishing returns. This mechanism ensures the model learns an optimal stopping policy, minimizing unnecessary iterations and maximizing inference efficiency in real-world scenarios.

\textbf{Tool-cost Penalty ($R_{cost}$):} Recognizing the computational cost of each interaction, we apply a linear penalty proportional to the sequence length. This encourages the policy to pursue the shortest path to high-fidelity segmentation, prioritizing efficiency without compromising terminal accuracy.

The terminal reward $R_{total}$ for each output $O_i$ is a composite metric that accounts for structural adherence, segmentation accuracy, and procedural efficiency:
\begin{equation}
\begin{aligned}
R_{total} = w_1 \cdot R_{fmt} + w_2 \cdot \text{clip}(R_{qual} \\+\lambda_1 R_{imp}-\lambda_2 R_{over} - \lambda_3 R_{cost}, 0, 1)
\end{aligned}
\end{equation}
where $w$ and $\lambda$ are weighted hyper-parameters, $\text{clip}(\cdot, a, b)$ limits the input to the range $[a, b]$. To optimize the interactive policy, we employ Group Relative Policy Optimization (GRPO \cite{GRPO}). The objective function $\mathcal{J}_{GRPO}(\theta)$ is defined by sampling a group of $G$ rollouts $\{O_1, O_2, \dots, O_G\}$ for each query $q$:
\begin{equation}
\begin{aligned}
\mathcal{J}_{GRPO}(\theta) = & \mathbb{E}_{q \sim \mathcal{D}, \{O_i\}_{i=1}^G \sim \pi_{\theta_{old}}} \bigg[ \frac{1}{G} \sum_{i=1}^G \min\Big( \\
\frac{\pi_\theta(O_i|q)}{\pi_{\theta_{old}}(O_i|q)} A_i,
& \text{clip} \left( \frac{\pi_\theta(O_i|q)}{\pi_{\theta_{old}}(O_i|q)}, 1-\epsilon, 1+\epsilon \right) A_i \Big) \bigg]
\end{aligned}
\end{equation}
where the advantage $A_i$ is computed by normalizing the multi-component reward $R_{total}$ within each group to capture the relative quality of the rollouts $O_i$:
\begin{equation}
A_i = \frac{R_{total, i} - \text{mean}(R_{total, 1}, \dots, R_{total, G})}{\text{std}(R_{total, 1}, \dots, R_{total, G})}
\end{equation}

By leveraging this relative advantage formulation, we ensure that the model learns to prioritize higher-reward trajectories. Through carefully designed RL training, the agent becomes capable of interpreting complex target descriptions and performing pixel-level reasoning via iterative tool invocations. Moreover, we specifically curate 9K high-quality samples for the RL training data to effectively guide this learning process and ensure strong policy convergence during the RL rollout. This selection focuses on challenging instances that necessitated 3–5 interaction rounds under both Box-to-Point and Sequential-Click paradigms, thereby exposing the model to complex decision-making scenarios and maximizing the efficiency of the optimization strategy.

\section{Experiments}

\subsection{Experiments Setting}

\noindent \textbf{Datasets.} We utilize a comprehensive collection of 21 open-source datasets spanning 6 modalities, including CT \cite{btcv,amos,FLARE,KiTS,LIDC}, MRI \cite{ACDC,amos,LGG}, X-Ray \cite{CXRMask,chowdhury2020can,CDD-CESM}, Ultrasound \cite{US,FH-PS-AOP}, Fundus \cite{refuge}, and Endoscopy \cite{NeoPolyp,polypgen}. All data formats are standardized following previous methods \cite{biomedparse,UniBiomed}. Details are available in the Appendix.

\textbf{Implementation Details.} In our experiments, the policy model is based on the Qwen3-VL-8B \cite{Qwen2.5-VL}, and the interactive segmentation models used in our work include general-purpose model SAM2.1-Base \cite{SAM2}, and medical domain-specific models MedSAM2 \cite{medsam2} and IMISNet \cite{IMISNet}. The SFT stage is implemented using Llama-Factory \cite{llamafactory} with a learning rate of 1e-5 and the batch size is 64. The RL stage is implemented by the Verl framework \cite{verl} with a learning rate of 1e-5, batch size of 8 and sampling number of 8, the maximum interaction turn is 5. All experiments are conducted on 8$\times$ NVIDIA H20 GPUs.

\subsection{Comparison Experiment}

\noindent \textbf{Baselines and Protocols.} In this section, we conduct a comprehensive evaluation of our MedSAM-Agent against existing state-of-the-art (SOTA) methods, including the interactive segmentations like general-purpose model SAM2~\cite{SAM2} and domain-specific models such as MedSAM2~\cite{medsam2} and IMISNet~\cite{IMISNet}. For these baselines, we report results using both \textit{Point} and \textit{Box} prompts. ``Point" results are generated using the center of the ground-truth mask, while ``Box" results utilize corresponding bounding boxes. The Box prompt performance is generally regarded as the empirical upper bound for single-turn interactive segmentation. Furthermore, we benchmark against SOTA MLLM-based methods, including four general-domain models (LISA~\cite{LISA}, GLAMM~\cite{GALLM}, HyperSeg~\cite{hyperseg}, and Seg-R1~\cite{Seg-R1}) and three specialized medical MLLMs (MedPLIB~\cite{MedPLIB}, UniBioMed~\cite{UniBiomed}, and Citrus-V~\cite{citrusV}). 

\begin{figure}[t!]
  \centering 
  \includegraphics[width=\linewidth]{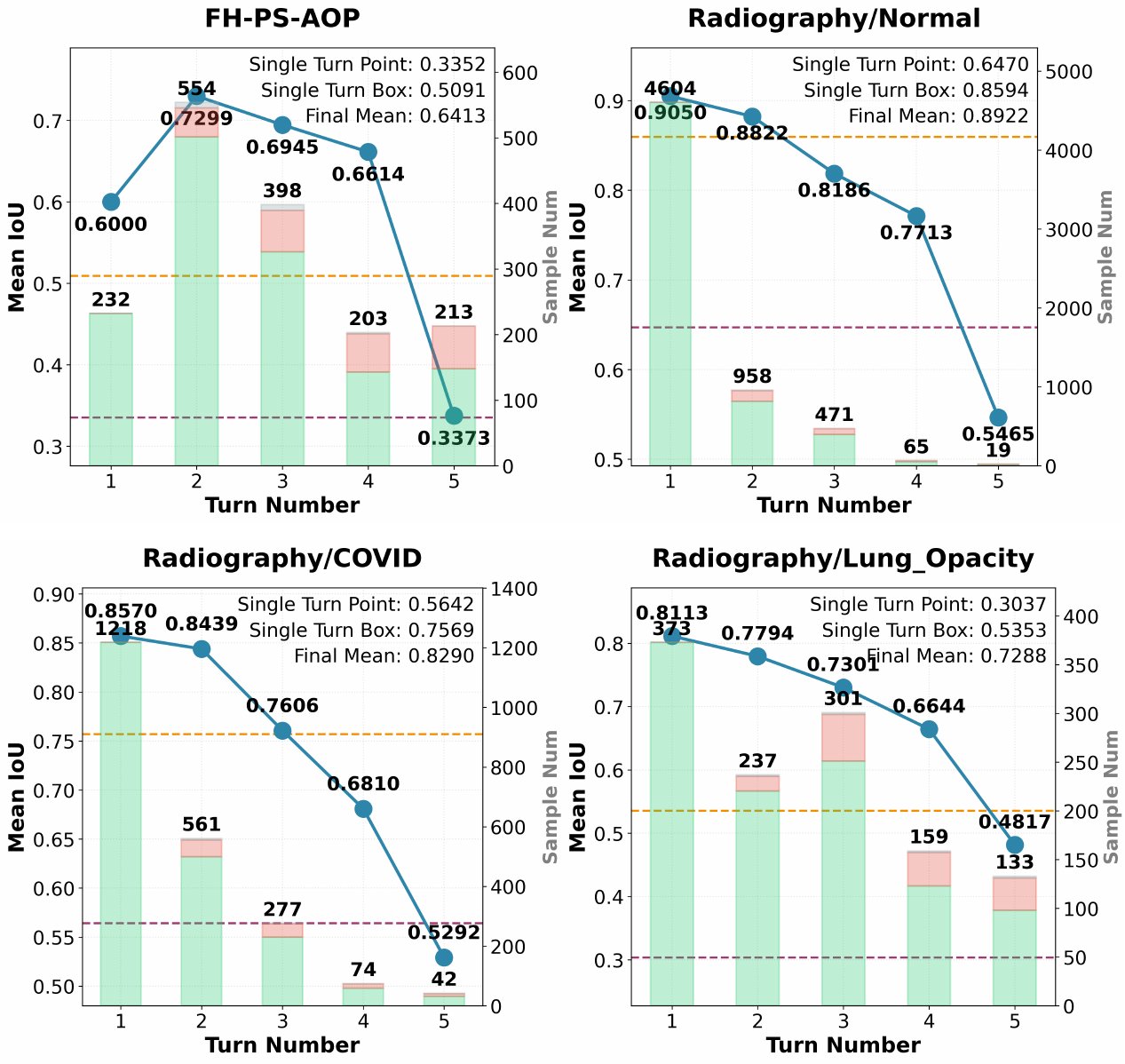}
\caption{\textbf{Analysis of multi-turn interaction.} The \textcolor[HTML]{F18F01}{orange} and \textcolor[HTML]{A23B72}{purple} lines represent the performance of the static Single-Turn Box and Single-Turn Point prompts, respectively, where inputs are derived from ground-truth bounding boxes and centroids. The \textcolor[HTML]{2E86AB}{blue} line plots the Mean IoU across successive interaction turns. The bar charts illustrate the distribution of sample outcomes at each turn, where \textcolor[HTML]{2ECC71}{green}, \textcolor[HTML]{E74C3C}{red}, and \textcolor[HTML]{95A5A6}{grey} segments denote the proportion of samples exhibiting improved, declined, or unchanged IoU. The segmentation tool is IMISNet~\cite{IMISNet}.}
\label{fig:fig3}
\end{figure} 

\noindent \textbf{Segmentation Performance.} Table \ref{tab:tab1} shows the quantitative comparison across 6 medical imaging modalities. We observe that general-purpose MLLMs, such as LISA \cite{LISA}, exhibit poor performance on these medical datasets. This suggests that the \texttt{<seg>} token-based paradigm suffers from limited generalization and faces significant hurdles when transferring to the specialized medical domain without extensive in-domain fine-tuning. In contrast, although Seg-R1 \cite{Seg-R1} was not specifically fine-tuned on medical domain data, it significantly outperforms other general-purpose MLLMs. This suggests that RL-based optimization inherently equips models with superior structural reasoning and out-of-distribution generalization for segmentation tasks, even when navigating unfamiliar medical modalities. Furthermore, specialized medical models such as UniBioMed and Citrus-V show marked improvements over general-purpose baselines, underscoring the vital importance of domain-specific medical knowledge. Building upon this, MedSAM-Agent further extends these gains by internalizing multi-turn interaction logic, consistently achieving high mask fidelity across all evaluated datasets. Details of the segmentation performance of 21 datasets are shown in the Appendix.

\begin{table}[t!]
\centering
\caption{\textbf{Performance comparison of various policy models and segmentation frameworks.} (Tr/Te denotes the segmentation models used during training and testing, respectively; `-' indicates zero-shot evaluation.}
\setlength{\tabcolsep}{1.8pt}
\begin{adjustbox}{width=0.49\textwidth,center} 
\begin{tabular}{lcccccc}
\toprule
\multirow{2}{*}{\textbf{Policy Model}} & \multirow{2}{*}{\textbf{Seg Model (Tr/Te)}} & \multicolumn{2}{c}{\textbf{BTCV}} & \multicolumn{2}{c}{\textbf{PolypGen}} \\
\cmidrule(lr){3-4} \cmidrule(lr){5-6}
& & Dice & IoU & Dice & IoU \\
\midrule
GPT-4o        & - / MedSAM2 & 0.122 & 0.104 & 0.549 & 0.481 \\
Qwen2.5-VL-7B & - / MedSAM2 & 0.117 & 0.098 & 0.389 & 0.331 \\
Qwen3-VL-8B   & - / MedSAM2 & 0.130 & 0.113 & 0.669 & 0.605 \\
Qwen3-VL-8B   & - / IMIS & 0.101 & 0.080 & 0.539 & 0.466 \\
\midrule
Qwen3-VL-8B   & IMIS / IMIS & \textbf{0.780} & \textbf{0.702} & 0.793 & 0.719 \\
Qwen3-VL-8B   & MedSAM2 / MedSAM2 & 0.773 & 0.690 & \textbf{0.809} & 0.735 \\
\midrule
Qwen3-VL-8B   & IMIS / SAM2 & 0.764 & 0.680 & 0.805 & 0.740 \\
Qwen3-VL-8B   & IMIS / MedSAM2 & 0.769 & 0.687 & 0.799 & 0.726 \\
Qwen3-VL-8B   & MedSAM2 / IMIS & 0.761 & 0.682 & 0.798 & 0.720 \\
Qwen3-VL-8B   & MedSAM2 / SAM2 & 0.763 & 0.677 & 0.808 & \textbf{0.739} \\

\bottomrule
\end{tabular}
\end{adjustbox}
\label{tab:model_comparison}
\end{table}

Moreover, we analyze the transformative power of the multi-turn strategic interaction of MedSAM-Agent. To demonstrate the efficacy of autonomous interaction, we compare our agent against static baseline paradigms, specifically Single-Turn Point and Single-Turn Box prompts where inputs are directly derived from ground-truth centroids and bounding boxes. As illustrated in Fig.~\ref{fig:fig3}, which details the performance trajectories on the FH-PS-AOP \cite{FH-PS-AOP} and Radiography \cite{chowdhury2020can} datasets, MedSAM-Agent outperforms these ideal single-turn baselines. Notably, the Final Mean IoU achieved by our agent exceeds the theoretical upper bound of static prompts across all evaluated modalities, proving that the agent has internalized a sophisticated refinement logic rather than simple tool invocation. The stacked bar charts further quantify this effectiveness: the predominance of green segments signifies that the vast majority of autonomous interactions result in substantial IoU gains. Furthermore, the diminishing sample counts across successive turns confirm the agent's ability to adaptively select the optimal interaction depth, effectively balancing segmentation precision with procedural efficiency.

\noindent \textbf{Zero-shot Tool Agnosticism.} To investigate the model's ability to generalize across various interactive segmentation backends, we conducted a cross-tool evaluation by separately synthesizing training trajectories using two distinct models, MedSAM2 and IMISNet, and subsequently testing the learned policies against alternative engines including SAM2. Our experiments reveal that while trajectories curated via MedSAM2 generally result in superior agent performance due to its higher fidelity in mask boundary delineations, the resulting agent exhibits exceptional zero-shot generalization regardless of the training backend. As summarized in Table \ref{tab:model_comparison}, an agent trained on one specific tool can be seamlessly interfaced with alternative backends at inference time with negligible performance degradation. This tool-agnostic capability suggests that the RL-based policy has successfully decoupled high-level strategic decision-making involving optimal point placement and autonomous stopping criteria from the low-level mask generation mechanics of specific tools. Consequently, MedSAM-Agent successfully internalizes a universal interaction logic, demonstrating its robust transferability and potential for integration into diverse clinical workflows independent of the underlying segmentation engine.

\subsection{Ablation Study}

\noindent \textbf{Effectiveness of Action Strategies.} We first investigate the impact of different interaction strategies on segmentation performance. As illustrated in Table \ref{tab:ablation}, the Hybrid strategy clearly outperforms any single-strategy approach. Compared to a purely Sequential-Click strategy, the use of the box-to-point strategy alone significantly improves the IoU from 0.623 to 0.649 while reducing the average interaction turns from 2.94 to 2.06. This demonstrates that the box strategy provides a superior spatial constraint that effectively defines target boundaries with fewer interactions. Most importantly, the Hybrid strategy achieves the highest performance among the SFT variants, reaching an IoU of 0.686. This success stems from the fact that the hybrid approach aligns with human intuition: it leverages the box to establish a reliable global context and uses subsequent points for precise local refinement. These results prove that combining box and point strategies is the most effective way to navigate complex medical lesions, as it minimizes redundant actions while maximizing segmentation fidelity.

\noindent \textbf{Analysis of Training Strategies.} We evaluate three training paradigms: direct RL from scratch, SFT only, and our combined SFT+RL approach (Table \ref{tab:ablation}). Direct RL exhibits a clear performance dichotomy across different modalities. In scenarios with prominent lesions and relatively simple backgrounds, such as X-ray and Endoscopy, direct RL can achieve competitive results as the agent easily identifies the target regions. However, it fails to achieve precise localization in more challenging modalities like CT and MRI, where the vast search space and low tissue contrast pose significant obstacles (Table.~\ref{tab:tab1}). This discrepancy highlights that while RL is adept at strategy optimization, it struggles with the initial grounding problem in complex medical contexts. In contrast, our SFT cold-start phase establishes essential anatomical grounding and domain knowledge. The subsequent RL stage then enables the model to generalize across tools and optimize for procedural efficiency. This two-stage pipeline ensures the agent possesses both the anatomical awareness required for medical precision and the strategic flexibility for efficient, multi-step interaction.

\noindent \textbf{Impact of Reward Design.} To evaluate the contribution of each component in our proposed reward framework, we conduct an ablation study on the process-aware rewards as shown in Table~\ref{tab:ablation}. Our results indicate that while the baseline model achieves basic segmentation, it often suffers from inefficient interaction trajectories and redundant operations. The inclusion of the $R_{imp}$ is critical for maintaining mask fidelity throughout the interaction; its removal results in a decline in Dice score from 0.794 to 0.772 and a drop in IoU to 0.688. The $R_{over}$ serves as a vital regularizer for the agent to determine the optimal timing for task completion. When $R_{over}$ is excluded, the agent fails to terminate the process at optimal states, leading to an increase in average interaction turns ($n_{turn}$) from 2.11 to 2.24. Similarly, the absence of the $R_{cost}$ confirms that our process-level supervision effectively discourages redundant strategies. Overall, the two-stage pipeline achieves the superior balance of precision and interaction turns, demonstrating that process-aware rewards successfully align the agent's behavior with both clinical accuracy and operational efficiency.

\begin{table}[t]
\centering
\caption{\textbf{Ablation study.} Metrics are average Dice and IoU, and the segmentation tool is MedSAM2 \cite{medsam2}.}
\begin{adjustbox}{width=0.48\textwidth,center}
\begin{tabular}{lccc}
\toprule
\textbf{Setting} & Dice & IoU & $n_{turn}$ \\ 
\midrule
Qwen3-VL-8B base   & 0.263  &  0.178   & 4.34     \\ 
\midrule
+ Cold-start SFT only (Hybrid) &  0.769 & 0.686 &  2.35  \\
- only Sequential-Click & 0.719 &  0.623   & 2.94  \\ 
- only Box-to-Point &   0.745 &  0.649   & \textbf{2.06}    \\
\midrule
+ RL only        & 0.678         & 0.593     & 2.23   \\
+ Cold-start SFT + RL (Full) & \textbf{0.794}  & \textbf{0.705}   & 2.11   \\
\midrule
- w/o $R_{imp}$ & 0.772   &  0.688       & 2.08      \\
- w/o $R_{over}$ &   0.785   & 0.693   & 2.24    \\
- w/o $R_{cost}$ &   0.788   & 0.696   & 2.26       \\
\bottomrule

\end{tabular}
\end{adjustbox}
\label{tab:ablation}
\end{table}


\section{Conclusion}
In this paper, we propose MedSAM-Agent, a novel framework that shifts the medical image segmentation paradigm from static classification to an autonomous, multi-step decision-making process. By addressing the limitations of current interactive models that depend on human guidance and automated MLLM approaches that lack iterative refinement, our method successfully bridges the gap between high-level reasoning and precise tool interaction. Through a hybrid prompting strategy and a two-stage training pipeline optimized by fine-grained process rewards, we enable the model to function as a self-refinement agent capable of achieving expert-level segmentation across diverse imaging modalities. Experimental results underscore the framework's superior performance and robust generalization, demonstrating its potential to reduce the workload of clinical professionals.

\bibliography{example_paper}

@String(CVPR= {IEEE Conf. Comput. Vis. Pattern Recog.})

@String(ICCV= {Int. Conf. Comput. Vis.})

@String(ECCV= {Eur. Conf. Comput. Vis.})

@String(AAAI = {AAAI})

@String(CVPR  = {CVPR})

@String(ICCV  = {ICCV})

@String(ECCV  = {ECCV})

@inproceedings{LISA,
  title={Lisa: Reasoning segmentation via large language model},
  author={Lai, Xin and Tian, Zhuotao and Chen, Yukang and Li, Yanwei and Yuan, Yuhui and Liu, Shu and Jia, Jiaya},
  booktitle={CVPR},
  pages={9579--9589},
  year={2024}
}

@article{UniBiomed,
  title={UniBiomed: A Universal Foundation Model for Grounded Biomedical Image Interpretation},
  author={Wu, Linshan and Nie, Yuxiang and He, Sunan and Zhuang, Jiaxin and Luo, Luyang and Mahboobani, Neeraj and Vardhanabhuti, Varut and Chan, Ronald Cheong Kin and Peng, Yifan and Rajpurkar, Pranav and others},
  journal={arXiv preprint arXiv:2504.21336},
  year={2025}
}

@article{MedSAM,
  title={Segment anything in medical images},
  author={Ma, Jun and He, Yuting and Li, Feifei and Han, Lin and You, Chenyu and Wang, Bo},
  journal={Nature Communications},
  volume={15},
  number={1},
  pages={654},
  year={2024},
  publisher={Nature Publishing Group UK London}
}

@article{SAM2,
  title={Sam 2: Segment anything in images and videos},
  author={Ravi, Nikhila and Gabeur, Valentin and Hu, Yuan-Ting and Hu, Ronghang and Ryali, Chaitanya and Ma, Tengyu and Khedr, Haitham and R{\"a}dle, Roman and Rolland, Chloe and Gustafson, Laura and others},
  journal={arXiv preprint arXiv:2408.00714},
  year={2024}
}

@article{SAM-R1,
  title={SAM-R1: Leveraging SAM for Reward Feedback in Multimodal Segmentation via Reinforcement Learning},
  author={Huang, Jiaqi and Xu, Zunnan and Zhou, Jun and Liu, Ting and Xiao, Yicheng and Ou, Mingwen and Ji, Bowen and Li, Xiu and Yuan, Kehong},
  journal={arXiv preprint arXiv:2505.22596},
  year={2025}
}

@article{Seg-R1,
  title={Seg-R1: Segmentation Can Be Surprisingly Simple with Reinforcement Learning},
  author={You, Zuyao and Wu, Zuxuan},
  journal={arXiv preprint arXiv:2506.22624},
  year={2025}
}

@inproceedings{SegAgent,
  title={Segagent: Exploring pixel understanding capabilities in mllms by imitating human annotator trajectories},
  author={Zhu, Muzhi and Tian, Yuzhuo and Chen, Hao and Zhou, Chunluan and Guo, Qingpei and Liu, Yang and Yang, Ming and Shen, Chunhua},
  booktitle={CVPR},
  pages={3686--3696},
  year={2025}
}

@inproceedings{llamafactory,
  title={LlamaFactory: Unified Efficient Fine-Tuning of 100+ Language Models},
  author={Zheng, Yaowei and Zhang, Richong and Zhang, Junhao and YeYanhan, YeYanhan and Luo, Zheyan},
  booktitle={ACL},
  pages={400--410},
  year={2024}
}

@article{GRPO,
  title={Deepseekmath: Pushing the limits of mathematical reasoning in open language models},
  author={Shao, Zhihong and Wang, Peiyi and Zhu, Qihao and Xu, Runxin and Song, Junxiao and Bi, Xiao and Zhang, Haowei and Zhang, Mingchuan and Li, YK and others},
  journal={arXiv preprint arXiv:2402.03300},
  year={2024}
}

@article{Qwen2.5-VL,
  title={Qwen2.5-VL Technical Report},
  author={Bai, Shuai and Chen, Keqin and Liu, Xuejing and Wang, Jialin and Ge, Wenbin and Song, Sibo and Dang, Kai and Wang, Peng and Wang, Shijie and Tang, Jun and Zhong, Humen and Zhu, Yuanzhi and Yang, Mingkun and Li, Zhaohai and Wan, Jianqiang and Wang, Pengfei and Ding, Wei and Fu, Zheren and Xu, Yiheng and Ye, Jiabo and Zhang, Xi and Xie, Tianbao and Cheng, Zesen and Zhang, Hang and Yang, Zhibo and Xu, Haiyang and Lin, Junyang},
  journal={arXiv preprint arXiv:2502.13923},
  year={2025}
}

@inproceedings{IMISNet,
  title={Interactive medical image segmentation: A benchmark dataset and baseline},
  author={Cheng, Junlong and Fu, Bin and Ye, Jin and Wang, Guoan and Li, Tianbin and Wang, Haoyu and Li, Ruoyu and Yao, He and Cheng, Junren and Li, JingWen and others},
  booktitle={Proceedings of the Computer Vision and Pattern Recognition Conference},
  pages={20841--20851},
  year={2025}
}

@article{Seg-Zero,
  title={Seg-zero: Reasoning-chain guided segmentation via cognitive reinforcement},
  author={Liu, Yuqi and Peng, Bohao and Zhong, Zhisheng and Yue, Zihao and Lu, Fanbin and Yu, Bei and Jia, Jiaya},
  journal={arXiv preprint arXiv:2503.06520},
  year={2025}
}

@article{VisionReasoner,
  title={VisionReasoner: Unified Visual Perception and Reasoning via Reinforcement Learning},
  author={Liu, Yuqi and Qu, Tianyuan and Zhong, Zhisheng and Peng, Bohao and Liu, Shu and Yu, Bei and Jia, Jiaya},
  journal={arXiv preprint arXiv:2505.12081},
  year={2025}
}

@article{pixelreasoner,
  title={Pixel reasoner: Incentivizing pixel-space reasoning with curiosity-driven reinforcement learning},
  author={Su, Alex and Wang, Haozhe and Ren, Weiming and Lin, Fangzhen and Chen, Wenhu},
  journal={arXiv preprint arXiv:2505.15966},
  year={2025}
}

@article{MedSAMAdapter,
  title={Medical sam adapter: Adapting segment anything model for medical image segmentation},
  author={Wu, Junde and Wang, Ziyue and Hong, Mingxuan and Ji, Wei and Fu, Huazhu and Xu, Yanwu and Xu, Min and Jin, Yueming},
  journal={Medical image analysis},
  volume={102},
  pages={103547},
  year={2025},
  publisher={Elsevier}
}

@inproceedings{SAM,
  title={Segment anything},
  author={Kirillov, Alexander and Mintun, Eric and Ravi, Nikhila and Mao, Hanzi and Rolland, Chloe and Gustafson, Laura and Xiao, Tete and Whitehead, Spencer and Berg, Alexander C and Lo, Wan-Yen and others},
  booktitle={CVPR},
  pages={4015--4026},
  year={2023}
}

@article{transunet,
  title={Transunet: Transformers make strong encoders for medical image segmentation},
  author={Chen, Jieneng and Lu, Yongyi and Yu, Qihang and Luo, Xiangde and Adeli, Ehsan and Wang, Yan and Lu, Le and Yuille, Alan L and Zhou, Yuyin},
  journal={arXiv preprint arXiv:2102.04306},
  year={2021}
}

@inproceedings{ukan,
  title={U-kan makes strong backbone for medical image segmentation and generation},
  author={Li, Chenxin and Liu, Xinyu and Li, Wuyang and Wang, Cheng and Liu, Hengyu and Liu, Yifan and Chen, Zhen and Yuan, Yixuan},
  booktitle={Proceedings of the AAAI Conference on Artificial Intelligence},
  volume={39},
  number={5},
  pages={4652--4660},
  year={2025}
}

@article{nnunet,
  title={nnU-Net: a self-configuring method for deep learning-based biomedical image segmentation},
  author={Isensee, Fabian and Jaeger, Paul F and Kohl, Simon AA and Petersen, Jens and Maier-Hein, Klaus H},
  journal={Nature methods},
  volume={18},
  number={2},
  pages={203--211},
  year={2021},
  publisher={Nature Publishing Group}
}

@inproceedings{medvlm-r1,
  title={Medvlm-r1: Incentivizing medical reasoning capability of vision-language models (vlms) via reinforcement learning},
  author={Pan, Jiazhen and Liu, Che and Wu, Junde and Liu, Fenglin and Zhu, Jiayuan and Li, Hongwei Bran and Chen, Chen and Ouyang, Cheng and Rueckert, Daniel},
  booktitle={MICCAI},
  pages={337--347},
  year={2025},
  organization={Springer}
}

@article{med-r1,
  title={Med-r1: Reinforcement learning for generalizable medical reasoning in vision-language models},
  author={Lai, Yuxiang and Zhong, Jike and Li, Ming and Zhao, Shitian and Yang, Xiaofeng},
  journal={arXiv preprint arXiv:2503.13939},
  year={2025}
}

@inproceedings{medground-r1,
  title={Medground-r1: Advancing medical image grounding via spatial-semantic rewarded group relative policy optimization},
  author={Xu, Huihui and Nie, Yuanpeng and Wang, Hualiang and Chen, Ying and Li, Wei and Ning, Junzhi and Liu, Lihao and Wang, Hongqiu and Zhu, Lei and Liu, Jiyao and others},
  booktitle={MICCAI},
  pages={391--401},
  year={2025},
  organization={Springer}
}

@article{Med-rlvr,
  title={Med-rlvr: Emerging medical reasoning from a 3b base model via reinforcement learning},
  author={Zhang, Sheng and Liu, Qianchu and Qin, Guanghui and Naumann, Tristan and Poon, Hoifung},
  journal={arXiv preprint arXiv:2502.19655},
  year={2025}
}

@article{VisualRFT,
  title={Visual-rft: Visual reinforcement fine-tuning},
  author={Liu, Ziyu and Sun, Zeyi and Zang, Yuhang and Dong, Xiaoyi and Cao, Yuhang and Duan, Haodong and Lin, Dahua and Wang, Jiaqi},
  journal={ICCV},
  year={2025}
}

@inproceedings{I-MedSAM,
  title={I-medsam: Implicit medical image segmentation with segment anything},
  author={Wei, Xiaobao and Cao, Jiajun and Jin, Yizhu and Lu, Ming and Wang, Guangyu and Zhang, Shanghang},
  booktitle={ECCV},
  pages={90--107},
  year={2024},
  organization={Springer}
}

@inproceedings{MedPLIB,
  title={Towards a multimodal large language model with pixel-level insight for biomedicine},
  author={Huang, Xiaoshuang and Shen, Lingdong and Liu, Jia and Shang, Fangxin and Li, Hongxiang and Huang, Haifeng and Yang, Yehui},
  booktitle={Proceedings of the AAAI Conference on Artificial Intelligence},
  volume={39},
  number={4},
  pages={3779--3787},
  year={2025}
}

@inproceedings{UniSeg,
  title={Uniseg: A prompt-driven universal segmentation model as well as a strong representation learner},
  author={Ye, Yiwen and Xie, Yutong and Zhang, Jianpeng and Chen, Ziyang and Xia, Yong},
  booktitle={MICCAI},
  pages={508--518},
  year={2023},
  organization={Springer}
}

@inproceedings{universeg,
  title={Universeg: Universal medical image segmentation},
  author={Butoi, Victor Ion and Ortiz, Jose Javier Gonzalez and Ma, Tianyu and Sabuncu, Mert R and Guttag, John and Dalca, Adrian V},
  booktitle={ICCV},
  pages={21438--21451},
  year={2023}
}

@inproceedings{EnhancingSAMMed,
  title={Enhancing SAM with Efficient Prompting and Preference Optimization for Semi-supervised Medical Image Segmentation},
  author={Konwer, Aishik and Yang, Zhijian and Bas, Erhan and Xiao, Cao and Prasanna, Prateek and Bhatia, Parminder and Kass-Hout, Taha},
  booktitle={CVPR},
  pages={20990--21000},
  year={2025}
}

@article{MedSegX,
  title={A generalist foundation model and database for open-world medical image segmentation},
  author={Zhang, Siqi and Zhang, Qizhe and Zhang, Shanghang and Liu, Xiaohong and Yue, Jingkun and Lu, Ming and Xu, Huihuan and Yao, Jiaxin and Wei, Xiaobao and Cao, Jiajun and others},
  journal={Nature Biomedical Engineering},
  pages={1--16},
  year={2025},
  publisher={Nature Publishing Group UK London}
}

@article{biomedparse,
  title={A foundation model for joint segmentation, detection and recognition of biomedical objects across nine modalities},
  author={Zhao, Theodore and Gu, Yu and Yang, Jianwei and Usuyama, Naoto and Lee, Ho Hin and Kiblawi, Sid and Naumann, Tristan and Gao, Jianfeng and Crabtree, Angela and Abel, Jacob and others},
  journal={Nature methods},
  volume={22},
  number={1},
  pages={166--176},
  year={2025},
  publisher={Nature Publishing Group US New York}
}

@inproceedings{unet,
  title={U-net: Convolutional networks for biomedical image segmentation},
  author={Ronneberger, Olaf and Fischer, Philipp and Brox, Thomas},
  booktitle={MICCAI},
  pages={234--241},
  year={2015},
  organization={Springer}
}

@article{unimedvl,
  title={Unimedvl: Unifying Medical Multimodal Understanding And Generation Through Observation-Knowledge-Analysis},
  author={Ning, Junzhi and Li, Wei and Tang, Cheng and Lin, Jiashi and Ma, Chenglong and Zhang, Chaoyang and Liu, Jiyao and Chen, Ying and Gao, Shujian and Liu, Lihao and others},
  journal={arXiv preprint arXiv:2510.15710},
  year={2025}
}

@article{DeepEyes,
  title={DeepEyes: Incentivizing" Thinking with Images" via Reinforcement Learning},
  author={Zheng, Ziwei and Yang, Michael and Hong, Jack and Zhao, Chenxiao and Xu, Guohai and Yang, Le and Shen, Chao and Yu, Xing},
  journal={arXiv preprint arXiv:2505.14362},
  year={2025}
}

@article{minio3,
  title={Mini-o3: Scaling up reasoning patterns and interaction turns for visual search},
  author={Lai, Xin and Li, Junyi and Li, Wei and Liu, Tao and Li, Tianjian and Zhao, Hengshuang},
  journal={arXiv preprint arXiv:2509.07969},
  year={2025}
}

@article{video-r1,
  title={Video-r1: Reinforcing video reasoning in mllms},
  author={Feng, Kaituo and Gong, Kaixiong and Li, Bohao and Guo, Zonghao and Wang, Yibing and Peng, Tianshuo and Wu, Junfei and Zhang, Xiaoying and Wang, Benyou and Yue, Xiangyu},
  journal={arXiv preprint arXiv:2503.21776},
  year={2025}
}

@article{thinkingwithvideos,
  title={Thinking with videos: Multimodal tool-augmented reinforcement learning for long video reasoning},
  author={Zhang, Haoji and Gu, Xin and Li, Jiawen and Ma, Chixiang and Bai, Sule and Zhang, Chubin and Zhang, Bowen and Zhou, Zhichao and He, Dongliang and Tang, Yansong},
  journal={arXiv preprint arXiv:2508.04416},
  year={2025}
}

@article{Lmm-r1,
  title={Lmm-r1: Empowering 3b lmms with strong reasoning abilities through two-stage rule-based rl},
  author={Peng, Yingzhe and Zhang, Gongrui and Zhang, Miaosen and You, Zhiyuan and Liu, Jie and Zhu, Qipeng and Yang, Kai and Xu, Xingzhong and Geng, Xin and Yang, Xu},
  journal={arXiv preprint arXiv:2503.07536},
  year={2025}
}

@article{vlm-r1,
  title={Vlm-r1: A stable and generalizable r1-style large vision-language model},
  author={Shen, Haozhan and Liu, Peng and Li, Jingcheng and Fang, Chunxin and Ma, Yibo and Liao, Jiajia and Shen, Qiaoli and Zhang, Zilun and Zhao, Kangjia and Zhang, Qianqian and others},
  journal={arXiv preprint arXiv:2504.07615},
  year={2025}
}

@inproceedings{shen2025zoomeye,
  title={Zoomeye: Enhancing multimodal llms with human-like zooming capabilities through tree-based image exploration},
  author={Shen, Haozhan and Zhao, Kangjia and Zhao, Tiancheng and Xu, Ruochen and Zhang, Zilun and Zhu, Mingwei and Yin, Jianwei},
  booktitle={EMNLP},
  pages={6613--6629},
  year={2025}
}

@misc{openai-o3,
  title = {Thinking with images},
  author = {OpenAI},
  year = {2025},
  url = {https://openai.com/index/thinking-with-images/},
}

@article{Lingshu,
  title={Lingshu: A Generalist Foundation Model for Unified Multimodal Medical Understanding and Reasoning},
  author={Xu, Weiwen and Chan, Hou Pong and Li, Long and Aljunied, Mahani and Yuan, Ruifeng and Wang, Jianyu and Xiao, Chenghao and Chen, Guizhen and Liu, Chaoqun and Li, Zhaodonghui and others},
  journal={arXiv preprint arXiv:2506.07044},
  year={2025}
}

@article{medthinkiter,
  title={Think Twice to See More: Iterative Visual Reasoning in Medical VLMs},
  author={Chen, Kaitao and Rui, Shaohao and Jiang, Yankai and Wu, Jiamin and Zheng, Qihao and Song, Chunfeng and Wang, Xiaosong and Zhou, Mu and Liu, Mianxin},
  journal={arXiv preprint arXiv:2510.10052},
  year={2025}
}

@article{OpenThinkIMG,
  title={Openthinkimg: Learning to think with images via visual tool reinforcement learning},
  author={Su, Zhaochen and Li, Linjie and Song, Mingyang and Hao, Yunzhuo and Yang, Zhengyuan and Zhang, Jun and Chen, Guanjie and Gu, Jiawei and Li, Juntao and Qu, Xiaoye and others},
  journal={arXiv preprint arXiv:2505.08617},
  year={2025}
}

@article{DAPO,
  title={Dapo: An open-source llm reinforcement learning system at scale},
  author={Yu, Qiying and Zhang, Zheng and Zhu, Ruofei and Yuan, Yufeng and Zuo, Xiaochen and Yue, Yu and Dai, Weinan and Fan, Tiantian and Liu, Gaohong and Liu, Lingjun and others},
  journal={arXiv preprint arXiv:2503.14476},
  year={2025}
}

@misc{GSPO,
      title={Group Sequence Policy Optimization}, 
      author={Chujie Zheng and Shixuan Liu and Mingze Li and Xiong-Hui Chen and Bowen Yu and Chang Gao and Kai Dang and Yuqiong Liu and Rui Men and An Yang and Jingren Zhou and Junyang Lin},
      year={2025},
      eprint={2507.18071},
      archivePrefix={arXiv},
      primaryClass={cs.LG},
      url={https://arxiv.org/abs/2507.18071}, 
}

@misc{PPO,
      title={Proximal Policy Optimization Algorithms}, 
      author={John Schulman and Filip Wolski and Prafulla Dhariwal and Alec Radford and Oleg Klimov},
      year={2017},
      eprint={1707.06347},
      archivePrefix={arXiv},
      primaryClass={cs.LG},
      url={https://arxiv.org/abs/1707.06347}, 
}

@misc{REINFORCE,
      title={Sample Efficient Reinforcement Learning with REINFORCE}, 
      author={Junzi Zhang and Jongho Kim and Brendan O'Donoghue and Stephen Boyd},
      year={2020},
      eprint={2010.11364},
      archivePrefix={arXiv},
      primaryClass={cs.LG},
      url={https://arxiv.org/abs/2010.11364}, 
}

@misc{thyme,
      title={Thyme: Think Beyond Images}, 
      author={Yi-Fan Zhang and Xingyu Lu and Shukang Yin and Chaoyou Fu and Wei Chen and Xiao Hu and Bin Wen and Kaiyu Jiang and Changyi Liu and Tianke Zhang and Haonan Fan and Kaibing Chen and Jiankang Chen and Haojie Ding and Kaiyu Tang and Zhang Zhang and Liang Wang and Fan Yang and Tingting Gao and Guorui Zhou},
      year={2025},
      eprint={2508.11630},
      archivePrefix={arXiv},
      primaryClass={cs.CV},
      url={https://arxiv.org/abs/2508.11630}, 
}

@misc{vtools-r1,
      title={VTool-R1: VLMs Learn to Think with Images via Reinforcement Learning on Multimodal Tool Use}, 
      author={Mingyuan Wu and Jingcheng Yang and Jize Jiang and Meitang Li and Kaizhuo Yan and Hanchao Yu and Minjia Zhang and Chengxiang Zhai and Klara Nahrstedt},
      year={2025},
      eprint={2505.19255},
      archivePrefix={arXiv},
      primaryClass={cs.LG},
      url={https://arxiv.org/abs/2505.19255}, 
}

@inproceedings{unet++,
  title={Unet++: A nested u-net architecture for medical image segmentation},
  author={Zhou, Zongwei and Rahman Siddiquee, Md Mahfuzur and Tajbakhsh, Nima and Liang, Jianming},
  booktitle={International workshop on deep learning in medical image analysis},
  pages={3--11},
  year={2018},
  organization={Springer}
}

@misc{pranet,
      title={PraNet: Parallel Reverse Attention Network for Polyp Segmentation}, 
      author={Deng-Ping Fan and Ge-Peng Ji and Tao Zhou and Geng Chen and Huazhu Fu and Jianbing Shen and Ling Shao},
      year={2020},
      eprint={2006.11392},
      archivePrefix={arXiv},
      primaryClass={eess.IV},
      url={https://arxiv.org/abs/2006.11392}, 
}

@misc{sam-med3d,
      title={SAM-Med3D: Towards General-purpose Segmentation Models for Volumetric Medical Images}, 
      author={Haoyu Wang and Sizheng Guo and Jin Ye and Zhongying Deng and Junlong Cheng and Tianbin Li and Jianpin Chen and Yanzhou Su and Ziyan Huang and Yiqing Shen and Bin Fu and Shaoting Zhang and Junjun He and Yu Qiao},
      year={2024},
      eprint={2310.15161},
      archivePrefix={arXiv},
      primaryClass={cs.CV},
      url={https://arxiv.org/abs/2310.15161}, 
}

@article{medsam2,
  title={Medsam2: Segment anything in 3d medical images and videos},
  author={Ma, Jun and Yang, Zongxin and Kim, Sumin and Chen, Bihui and Baharoon, Mohammed and Fallahpour, Adibvafa and Asakereh, Reza and Lyu, Hongwei and Wang, Bo},
  journal={arXiv preprint arXiv:2504.03600},
  year={2025}
}

@article{medreasoner,
  title={MedReasoner: Reinforcement Learning Drives Reasoning Grounding from Clinical Thought to Pixel-Level Precision},
  author={Yan, Zhonghao and Diao, Muxi and Yang, Yuxuan and Xu, Jiayuan and Zhang, Kaizhou and Jing, Ruoyan and Yang, Lele and Liu, Yanxi and Liang, Kongming and Ma, Zhanyu},
  journal={arXiv preprint arXiv:2508.08177},
  year={2025}
}

@article{medseg-r,
  title={MedSeg-R: Reasoning Segmentation in Medical Images with Multimodal Large Language Models},
  author={Huang, Yu and Peng, Zelin and Zhao, Yichen and Yang, Piao and Yang, Xiaokang and Shen, Wei},
  journal={arXiv preprint arXiv:2506.10465},
  year={2025}
}

@misc{Huatuo-Vision,
      title={HuatuoGPT-Vision, Towards Injecting Medical Visual Knowledge into Multimodal LLMs at Scale}, 
      author={Junying Chen and Chi Gui and Ruyi Ouyang and Anningzhe Gao and Shunian Chen and Guiming Hardy Chen and Xidong Wang and Ruifei Zhang and Zhenyang Cai and Ke Ji and Guangjun Yu and Xiang Wan and Benyou Wang},
      year={2024},
      eprint={2406.19280},
      archivePrefix={arXiv},
      primaryClass={cs.CV},
}

@article{braingpt,
  title={Towards a holistic framework for multimodal LLM in 3D brain CT radiology report generation},
  author={Li, Cheng-Yi and Chang, Kao-Jung and Yang, Cheng-Fu and Wu, Hsin-Yu and Chen, Wenting and Bansal, Hritik and Chen, Ling and Yang, Yi-Ping and Chen, Yu-Chun and Chen, Shih-Pin and others},
  journal={Nature Communications},
  volume={16},
  number={1},
  pages={2258},
  year={2025},
  publisher={Nature Publishing Group UK London}
}

@misc{endobench,
      title={EndoBench: A Comprehensive Evaluation of Multi-Modal Large Language Models for Endoscopy Analysis}, 
      author={Shengyuan Liu and Boyun Zheng and Wenting Chen and Zhihao Peng and Zhenfei Yin and Jing Shao and Jiancong Hu and Yixuan Yuan},
      year={2025},
      eprint={2505.23601},
      archivePrefix={arXiv},
}

@inproceedings{omnimedvqa,
  title={Omnimedvqa: A new large-scale comprehensive evaluation benchmark for medical lvlm},
  author={Hu, Yutao and Li, Tianbin and Lu, Quanfeng and Shao, Wenqi and He, Junjun and Qiao, Yu and Luo, Ping},
  booktitle={Proceedings of the IEEE/CVF Conference on Computer Vision and Pattern Recognition},
  pages={22170--22183},
  year={2024}
}

@article{gmaimmbench,
  title={Gmai-mmbench: A comprehensive multimodal evaluation benchmark towards general medical ai},
  author={Ye, Jin and Wang, Guoan and Li, Yanjun and Deng, Zhongying and Li, Wei and Li, Tianbin and Duan, Haodong and Huang, Ziyan and Su, Yanzhou and Wang, Benyou and others},
  journal={Advances in Neural Information Processing Systems},
  volume={37},
  pages={94327--94427},
  year={2024}
}

@article{llavarad,
  title={A clinically accessible small multimodal radiology model and evaluation metric for chest X-ray findings},
  author={Zambrano Chaves, Juan Manuel and Huang, Shih-Cheng and Xu, Yanbo and Xu, Hanwen and Usuyama, Naoto and Zhang, Sheng and Wang, Fei and Xie, Yujia and Khademi, Mahmoud and Yang, Ziyi and others},
  journal={Nature Communications},
  volume={16},
  number={1},
  pages={3108},
  year={2025},
  publisher={Nature Publishing Group UK London}
}

@inproceedings{chen_reportgen,
  title={Fine-grained image-text alignment in medical imaging enables explainable cyclic image-report generation},
  author={Chen, Wenting and Shen, Linlin and Lin, Jingyang and Luo, Jiebo and Li, Xiang and Yuan, Yixuan},
  booktitle={Proceedings of the 62nd Annual Meeting of the Association for Computational Linguistics (Volume 1: Long Papers)},
  pages={9494--9509},
  year={2024}
}

@inproceedings{GALLM,
  title={Glamm: Pixel grounding large multimodal model},
  author={Rasheed, Hanoona and Maaz, Muhammad and Shaji, Sahal and Shaker, Abdelrahman and Khan, Salman and Cholakkal, Hisham and Anwer, Rao M and Xing, Eric and Yang, Ming-Hsuan and Khan, Fahad S},
  booktitle={CVPR},
  pages={13009--13018},
  year={2024}
}

@article{verl,
  title   = {HybridFlow: A Flexible and Efficient RLHF Framework},
  author  = {Guangming Sheng and Chi Zhang and Zilingfeng Ye and Xibin Wu and Wang Zhang and Ru Zhang and Yanghua Peng and Haibin Lin and Chuan Wu},
  year    = {2024},
  journal = {arXiv preprint arXiv: 2409.19256}
}

@article{Ophiuchus,
  title={Incentivizing Tool-augmented Thinking with Images for Medical Image Analysis},
  author={Jiang, Yankai and Zhang, Yujie and Zhang, Peng and Li, Yichen and Chen, Jintai and Shi, Xiaoming and Zhen, Shihui},
  journal={arXiv preprint arXiv:2512.14157},
  year={2025}
}

@article{mmedagent,
  title={Mmedagent: Learning to use medical tools with multi-modal agent},
  author={Li, Binxu and Yan, Tiankai and Pan, Yuanting and Luo, Jie and Ji, Ruiyang and Ding, Jiayuan and Xu, Zhe and Liu, Shilong and Dong, Haoyu and Lin, Zihao and others},
  journal={arXiv preprint arXiv:2407.02483},
  year={2024}
}

@article{medagent-pro,
  title={MedAgent-Pro: Towards Evidence-Based Multi-Modal Medical Diagnosis via Reasoning Agentic Workflow},
  author={Wang, Ziyue and Wu, Junde and Cai, Linghan and Low, Chang Han and Yang, Xihong and Li, Qiaxuan and Jin, Yueming},
  journal={arXiv preprint arXiv:2503.18968},
  year={2025}
}

@article{polypgen,
  title={A multi-centre polyp detection and segmentation dataset for generalisability assessment},
  author={Ali, Sharib and Jha, Debesh and Ghatwary, Noha and Realdon, Stefano and Cannizzaro, Renato and Salem, Osama E and Lamarque, Dominique and Daul, Christian and Riegler, Michael A and Anonsen, Kim V and others},
  journal={Scientific Data},
  volume={10},
  number={1},
  pages={75},
  year={2023}
}

@article{citrusV,
  title={Citrus-V: Advancing Medical Foundation Models with Unified Medical Image Grounding for Clinical Reasoning},
  author={Wang, Guoxin and Zhao, Jun and Liu, Xinyi and Liu, Yanbo and Cao, Xuyang and Li, Chao and Liu, Zhuoyun and Sun, Qintian and Zhou, Fangru and Xing, Haoqiang and others},
  journal={arXiv preprint arXiv:2509.19090},
  year={2025}
}

@misc{hyperseg,
      title={HyperSeg: Towards Universal Visual Segmentation with Large Language Model}, 
      author={Cong Wei and Yujie Zhong and Haoxian Tan and Yong Liu and Zheng Zhao and Jie Hu and Yujiu Yang},
      year={2024},
      eprint={2411.17606},
      archivePrefix={arXiv},
      primaryClass={cs.CV},
      url={https://arxiv.org/abs/2411.17606}, 
}

@inproceedings{aura,
  title={Aura: A multi-modal medical agent for understanding, reasoning and annotation},
  author={Fathi, Nima and Kumar, Amar and Arbel, Tal},
  booktitle={International Workshop on Agentic AI for Medicine},
  pages={105--114},
  year={2025},
  organization={Springer}
}

@article{ibisagent,
  title={IBISAgent: Reinforcing Pixel-Level Visual Reasoning in MLLMs for Universal Biomedical Object Referring and Segmentation},
  author={Jiang, Yankai and Li, Qiaoru and Xu, Binlu and Sun, Haoran and Ding, Chao and Dong, Junting and Cai, Yuxiang and Zhang, Xuhong and Yin, Jianwei},
  journal={arXiv preprint arXiv:2601.03054},
  year={2026}
}

@article{pathbench,
  title={PathBench: A comprehensive comparison benchmark for pathology foundation models towards precision oncology},
  author={Ma, Jiabo and Xu, Yingxue and Zhou, Fengtao and Wang, Yihui and Jin, Cheng and Guo, Zhengrui and Wu, Jianfeng and Tang, On Ki and Zhou, Huajun and Wang, Xi and others},
  journal={arXiv preprint arXiv:2505.20202},
  year={2025}
}

@article{dermogpt,
  title={DermoGPT: Open Weights and Open Data for Morphology-Grounded Dermatological Reasoning MLLMs},
  author={Ru, Jinghan and Yan, Siyuan and Yin, Yuguo and Zou, Yuexian and Ge, Zongyuan},
  journal={arXiv preprint arXiv:2601.01868},
  year={2026}
}

@article{sglang,
  title={Sglang: Efficient execution of structured language model programs},
  author={Zheng, Lianmin and Yin, Liangsheng and Xie, Zhiqiang and Sun, Chuyue Livia and Huang, Jeff and Yu, Cody Hao and Cao, Shiyi and Kozyrakis, Christos and Stoica, Ion and Gonzalez, Joseph E and others},
  journal={Advances in neural information processing systems},
  volume={37},
  pages={62557--62583},
  year={2024}
}

@article{KiTS,
  title={The kits21 challenge: Automatic segmentation of kidneys, renal tumors, and renal cysts in corticomedullary-phase ct},
  author={Heller, Nicholas and Isensee, Fabian and Trofimova, Dasha and Tejpaul, Resha and Zhao, Zhongchen and Chen, Huai and Wang, Lisheng and Golts, Alex and Khapun, Daniel and Shats, Daniel and others},
  journal={arXiv preprint arXiv:2307.01984},
  year={2023}
}

@inproceedings{btcv,
  title={Miccai multi-atlas labeling beyond the cranial vault--workshop and challenge},
  author={Landman, Bennett and Xu, Zhoubing and Igelsias, J and Styner, Martin and Langerak, T and Klein, Arno},
  booktitle={Proc. MICCAI Multi-Atlas Labeling Beyond Cranial Vault—Workshop Challenge},
  volume={5},
  pages={12},
  year={2015}
}

@article{amos,
  title={Amos: A large-scale abdominal multi-organ benchmark for versatile medical image segmentation},
  author={Ji, Yuanfeng and Bai, Haotian and Ge, Chongjian and Yang, Jie and Zhu, Ye and Zhang, Ruimao and Li, Zhen and Zhanng, Lingyan and Ma, Wanling and Wan, Xiang and others},
  journal={Advances in Neural Information Processing Systems},
  volume={35},
  pages={36722--36732},
  year={2022}
}

@article{word,
  title={WORD: A large scale dataset, benchmark and clinical applicable study for abdominal organ segmentation from CT image},
  author={Luo, Xiangde and Liao, Wenjun and Xiao, Jianghong and Chen, Jieneng and Song, Tao and Zhang, Xiaofan and Li, Kang and Metaxas, Dimitris N and Wang, Guotai and Zhang, Shaoting},
  journal={Medical Image Analysis},
  volume={82},
  pages={102642},
  year={2022},
  publisher={Elsevier}
}

@article{FLARE,
  title={Fast and low-GPU-memory abdomen CT organ segmentation: the flare challenge},
  author={Ma, Jun and Zhang, Yao and Gu, Song and An, Xingle and Wang, Zhihe and Ge, Cheng and Wang, Congcong and Zhang, Fan and Wang, Yu and Xu, Yinan and others},
  journal={Medical Image Analysis},
  volume={82},
  pages={102616},
  year={2022},
  publisher={Elsevier}
}

@article{LIDC,
  title={The lung image database consortium (LIDC) and image database resource initiative (IDRI): a completed reference database of lung nodules on CT scans},
  author={Armato III, Samuel G and McLennan, Geoffrey and Bidaut, Luc and McNitt-Gray, Michael F and Meyer, Charles R and Reeves, Anthony P and Zhao, Binsheng and Aberle, Denise R and Henschke, Claudia I and Hoffman, Eric A and others},
  journal={Medical physics},
  volume={38},
  number={2},
  pages={915--931},
  year={2011},
  publisher={Wiley Online Library}
}

@article{total,
    author = {Wasserthal, Jakob and Breit, Hanns-Christian and Meyer, Manfred T. and Pradella, Maurice and Hinck, Daniel and Sauter, Alexander W. and Heye, Tobias and Boll, Daniel T. and Cyriac, Joshy and Yang, Shan and Bach, Michael and Segeroth, Martin},
    title = {TotalSegmentator: Robust Segmentation of 104 Anatomic Structures in CT Images},
    journal = {Radiology: Artificial Intelligence},
    volume = {5},
    number = {5},
    pages = {e230024},
    year = {2023}
}

@article{3dsam,
  title={3dsam-adapter: Holistic adaptation of sam from 2d to 3d for promptable tumor segmentation},
  author={Gong, Shizhan and Zhong, Yuan and Ma, Wenao and Li, Jinpeng and Wang, Zhao and Zhang, Jingyang and Heng, Pheng-Ann and Dou, Qi},
  journal={Medical Image Analysis},
  volume={98},
  pages={103324},
  year={2024},
  publisher={Elsevier}
}

@inproceedings{MedRegA,
  title={Interpretable bilingual multimodal large language model for diverse biomedical tasks},
  author={Wang, Lehan and Wang, Haonan and Yang, Honglong and Mao, Jiaji and Yang, Zehong and Shen, Jun and Li, Xiaomeng},
  booktitle={International Conference on Learning Representations},
  pages={},
  year={2025}
}

@article{ACDC,
  title={Deep learning techniques for automatic MRI cardiac multi-structures segmentation and diagnosis: is the problem solved?},
  author={Bernard, Olivier and Lalande, Alain and Zotti, Clement and Cervenansky, Frederick and Yang, Xin and Heng, Pheng-Ann and Cetin, Irem and Lekadir, Karim and Camara, Oscar and Ballester, Miguel Angel Gonzalez and others},
  journal={IEEE Transactions on Medical Imaging},
  volume={37},
  number={11},
  pages={2514--2525},
  year={2018},
  publisher={ieee}
}

@article{CXRMask,
  title={Two public chest X-ray datasets for computer-aided screening of pulmonary diseases},
  author={Jaeger, Stefan and Candemir, Sema and Antani, Sameer and W{\'a}ng, Y{\`\i}-Xi{\'a}ng J and Lu, Pu-Xuan and Thoma, George},
  journal={Quantitative imaging in medicine and surgery},
  volume={4},
  number={6},
  pages={475},
  year={2014}
}

@article{CDD-CESM,
  title={Categorized digital database for low energy and subtracted contrast enhanced spectral mammography images},
  author={Khaled, R and others},
  journal={The Cancer Imaging Archive},
  year={2021}
}

@article{chowdhury2020can,
  title={Can AI help in screening viral and COVID-19 pneumonia?},
  author={Chowdhury, Muhammad EH and Rahman, Tawsifur and Khandakar, Amith and Mazhar, Rashid and Kadir, Muhammad Abdul and Mahbub, Zaid Bin and Islam, Khandakar Reajul and Khan, Muhammad Salman and Iqbal, Atif and Al Emadi, Nasser and others},
  journal={IEEE Access},
  volume={8},
  pages={132665--132676},
  year={2020},
  publisher={IEEE}
}

@article{US,
  title={Improving realism in patient-specific abdominal ultrasound simulation using CycleGANs},
  author={Vitale, Santiago and Orlando, Jos{\'e} Ignacio and Iarussi, Emmanuel and Larrabide, Ignacio},
  journal={International Journal of Computer Assisted Radiology and Surgery},
  volume={15},
  number={2},
  pages={183--192},
  year={2020},
  publisher={Springer}
}

@article{LGG,
  title={Association of genomic subtypes of lower-grade gliomas with shape features automatically extracted by a deep learning algorithm},
  author={Buda, Mateusz and Saha, Ashirbani and Mazurowski, Maciej A},
  journal={Computers in Biology and Medicine},
  volume={109},
  pages={218--225},
  year={2019},
  publisher={Elsevier}
}

@misc{FH-PS-AOP,
  title={Pubic symphysis-fetal head segmentation and angle of progression},
  author={Jieyun, Bai and ZhanHong, Ou},
  year={2024}
}

@article{refuge,
  title={Refuge challenge: A unified framework for evaluating automated methods for glaucoma assessment from fundus photographs},
  author={Orlando, Jos{\'e} Ignacio and Fu, Huazhu and Breda, Jo{\~a}o Barbosa and Van Keer, Karel and Bathula, Deepti R and Diaz-Pinto, Andr{\'e}s and Fang, Ruogu and Heng, Pheng-Ann and Kim, Jeyoung and Lee, JoonHo and others},
  journal={Medical Image Analysis},
  volume={59},
  pages={101570},
  year={2020},
  publisher={Elsevier}
}

@inproceedings{NeoPolyp,
  title={Neounet: Towards accurate colon polyp segmentation and neoplasm detection},
  author={Ngoc Lan, Phan and An, Nguyen Sy and Hang, Dao Viet and Long, Dao Van and Trung, Tran Quang and Thuy, Nguyen Thi and Sang, Dinh Viet},
  booktitle={Advances in visual computing: 16th international symposium, ISVC 2021, virtual event, October 4-6, 2021, proceedings, part II},
  pages={15--28},
  year={2021},
  organization={Springer}
}

@article{pathvqa,
    title={PathVQA: 30000+ Questions for Medical Visual Question Answering},
    author={He, Xuehai and Zhang, Yichen and Mou, Luntian and Xing, Eric and Xie, Pengtao},
    journal={arXiv preprint arXiv:2003.10286},
    year={2020}
}

@inproceedings{slake,
  title={Slake: A semantically-labeled knowledge-enhanced dataset for medical visual question answering},
  author={Liu, Bo and Zhan, Li-Ming and Xu, Li and Ma, Lin and Yang, Yan and Wu, Xiao-Ming},
  booktitle={2021 IEEE 18th international symposium on biomedical imaging (ISBI)},
  pages={1650--1654},
  year={2021},
  organization={IEEE}
}

@article{VQA-RAD,
  title={A dataset of clinically generated visual questions and answers about radiology images},
  author={Lau, Jason J and Gayen, Soumya and Ben Abacha, Asma and Demner-Fushman, Dina},
  journal={Scientific data},
  volume={5},
  number={1},
  pages={1--10},
  year={2018},
  publisher={Nature Publishing Group}
}

@inproceedings{clip,
  title={Learning transferable visual models from natural language supervision},
  author={Radford, Alec and Kim, Jong Wook and Hallacy, Chris and Ramesh, Aditya and Goh, Gabriel and Agarwal, Sandhini and Sastry, Girish and Askell, Amanda and Mishkin, Pamela and Clark, Jack and others},
  booktitle={International Conference on Machine Learning},
  pages={8748--8763},
  year={2021},
  organization={PmLR}
}

@article{hulumed,
  title={Hulu-Med: A Transparent Generalist Model towards Holistic Medical Vision-Language Understanding},
  author={Jiang, Songtao and Wang, Yuan and Song, Sibo and Hu, Tianxiang and Zhou, Chenyi and Pu, Bin and Zhang, Yan and Yang, Zhibo and Feng, Yang and Zhou, Joey Tianyi and others},
  journal={arXiv preprint arXiv:2510.08668},
  year={2025}
}

@article{medgemma,
  title={Medgemma technical report},
  author={Sellergren, Andrew and Kazemzadeh, Sahar and Jaroensri, Tiam and Kiraly, Atilla and Traverse, Madeleine and Kohlberger, Timo and Xu, Shawn and Jamil, Fayaz and Hughes, C{\'\i}an and Lau, Charles and others},
  journal={arXiv preprint arXiv:2507.05201},
  year={2025}
}

@inproceedings{nath2025vila,
  title={Vila-m3: Enhancing vision-language models with medical expert knowledge},
  author={Nath, Vishwesh and Li, Wenqi and Yang, Dong and Myronenko, Andriy and Zheng, Mingxin and Lu, Yao and Liu, Zhijian and Yin, Hongxu and Law, Yee Man and Tang, Yucheng and others},
  booktitle={Proceedings of the Computer Vision and Pattern Recognition Conference},
  pages={14788--14798},
  year={2025}
}

@misc{reasoning_survey,
      title={Medical Reasoning in the Era of LLMs: A Systematic Review of Enhancement Techniques and Applications}, 
      author={Wenxuan Wang and Zizhan Ma and Meidan Ding and Shiyi Zheng and Shengyuan Liu and Jie Liu and Jiaming Ji and Wenting Chen and Xiang Li and Linlin Shen and Yixuan Yuan},
      year={2025},
      eprint={2508.00669},
      archivePrefix={arXiv},
      primaryClass={cs.CL},
      url={https://arxiv.org/abs/2508.00669}, 
}
\bibliographystyle{icml2026}

\newpage
\appendix
\onecolumn

\begin{center}
    \section*{{\huge Supplementary}}
\end{center}

\section{Datasets}
In our experiments, we utilize a comprehensive collection of 21 open-source datasets spanning 6 modalities, including CT (FLARE22 \cite{FLARE}, KiTS \cite{KiTS}, LIDC-IDRI \cite{LIDC}, BTCV \cite{btcv}, AMOS-CT \cite{amos}, and WORD \cite{word}), MRI (ACDC \cite{ACDC}, LGG \cite{LGG}, and AMOS-MRI \cite{amos}), X-Ray (CXRMask \cite{CXRMask}, Radiography series \cite{chowdhury2020can}, and CDD-CESM \cite{CDD-CESM}), Ultrasound (BreastUS \cite{US}, LiverUS \cite{US}, and FH-PS-AOP \cite{FH-PS-AOP}), Fundus (REFUGE \cite{refuge}), and Endoscopy (NeoPolyp \cite{NeoPolyp} and PolypGen \cite{polypgen}). All data formats are standardized following previous methods \cite{biomedparse,UniBiomed}.

\begin{table}[htbp]
\centering
\caption{Descriptions of datasets used in this work, including modalities, regions of interest, and the number of triplets (image-text-label).}
\begin{tabular}{llll}
\toprule
Dataset & Modality & Regions of interest & Number \\
\midrule
FLARE22 \cite{FLARE} & CT & Abdomen organs & 26,802 \\
KiTS \cite{KiTS} & CT & Kidney \& Kidney Tumor & 44,557 \\
LIDC-IDRI \cite{LIDC} & CT & Lung nodule & 9,122 \\
BTCV \cite{btcv} & CT & Abdomen organs & 12,176 \\
AMOS-CT \cite{amos} & CT & Abdomen organs & 138,371 \\
WORD \cite{word} & CT & Abdomen organs & 58,898 \\
ACDC \cite{ACDC} & MRI & Heart & 7,666 \\
LGG \cite{LGG} & MRI & Brain Tumor & 2,542 \\
AMOS-MRI \cite{amos} & MRI & Abdomen & 52,625 \\
CXRMask \cite{CXRMask} & X-Ray & Chest & 1,698 \\
Radiography-Lung-opacity \cite{chowdhury2020can} & X-Ray & Chest & 6,012 \\
Radiography-Normal \cite{chowdhury2020can} & X-Ray & Chest & 30,574 \\
Radiography-Viral-Pneumonia \cite{chowdhury2020can} & X-Ray & Chest & 1,345 \\
Radiography-COVID \cite{chowdhury2020can} & X-Ray & Chest & 10,844 \\
CDD-CESM \cite{CDD-CESM} & X-Ray & Breast lesion & 1,233 \\
BreastUS \cite{US} & Ultrasound & Breast lesion & 1,294 \\
LiverUS \cite{US} & Ultrasound & Liver & 39 \\
FH-PS-AOP \cite{FH-PS-AOP} & Ultrasound & Transperineal & 8,000 \\
REFUGE \cite{refuge} & Fundus & Retinal & 2,400 \\
NeoPolyp \cite{NeoPolyp} & Endoscopy & Colon polyp & 2,050 \\
PolypGen \cite{polypgen} & Endoscopy & Colon polyp & 1,411 \\
\midrule
Total & 6 Modalities & & 419,659\\
\bottomrule
\end{tabular}
\end{table}

\section{Implementation Details}

\subsection{Trajectory Construction}
To train our model for a multi-step medical image segmentation agent, we require a dataset of expert-like interaction trajectories. We employed an automated algorithm to generate these trajectories by simulating the sequential refinement process an expert annotator would perform. While existing methods \cite{SegAgent, ibisagent} rely on a rigid point-wise simulation driven by pixel-level discrepancies, they typically restrict the action space to sequential clicks. Such a limitation fails to capture the multi-modal nature of human workflows, where practitioners often initiate segmentation by defining a Bounding Box before refining with precise clicks. To better align with human intuition, we propose a hybrid prompting strategy, as detailed in Algorithm \ref{alg:hybrid_prompting}. This strategy encompasses both Box-to-Point and Sequential-Click paradigms to generate diverse and realistic interaction trajectories.

\begin{algorithm}[H]
\caption{Hybrid Prompting Strategy for Trajectory Generation}
\label{alg:hybrid_prompting}
\begin{algorithmic}[1]
\REQUIRE Image $I$, ground truth mask $M_{\text{target}}$, state $s$, segmentation model $f_\theta$, max clicks $K$, gain threshold $\tau$, \\
max retries $N$

\ENSURE Trajectory $\mathcal{T} = \{(a_0, M_0), (a_1, M_1), \ldots, (a_T, M_T)\}$\STATE $\text{paradigm} \gets \{\text{``Box-to-Point"}, \text{``Sequential-Click"}\}
$\STATE $\mathcal{T} \gets \emptyset$\STATE \COMMENT{Initialization at $t=0$ using either Bounding Box or Centroid Click}\IF{$\text{paradigm} = \text{``Box-to-Point"}$}\STATE $a_0 \gets \text{ExtractBoundingBox}(M_{\text{target}}) + \epsilon, \text{ where } \epsilon \sim \mathcal{U}(-\delta, \delta)$\ELSE\STATE $a_0 \gets \frac{1}{|M_{\text{target}}|} \sum_{(x,y) \in M_{\text{target}}} (x, y) + \epsilon$\ENDIF\STATE $M_0 \gets f_\theta(I, a_0)$\STATE $\mathcal{T} \gets \mathcal{T} \cup \{(a_0, M_0)\}$\FOR{$t = 1$ \TO $K$}\STATE $\text{FN}_t \gets M_{\text{target}} \setminus M_{t-1}, \quad \text{FP}_t \gets M_{t-1} \setminus M_{\text{target}}$\IF{$|\text{FN}_t| = 0$ \AND $|\text{FP}_t| = 0$} \STATE \textbf{break} \ENDIF\STATE \COMMENT{Error-driven refinement with retry mechanism}
\STATE $\text{success} \gets \text{False}$
\FOR{$\text{trial} = 1$ \TO $N$}
    \IF{$|\text{FN}_t| > |\text{FP}_t|$}
        \STATE $a_t \gets \arg\max_{p \in \text{FN}_t} \mathcal{D}(p)$ \quad (Positive Click)
    \ELSE
        \STATE $a_t \gets \arg\max_{p \in \text{FP}_t} \mathcal{D}(p)$ \quad (Negative Click)
    \ENDIF
    
    \STATE $M_t \gets f_\theta(I, s_{t-1})$
    \STATE $\Delta \text{IoU}_t \gets \text{IoU}(M_t, M_{\text{target}}) - \text{IoU}(M_{t-1}, M_{\text{target}})$
    
    \IF{$\Delta \text{IoU}_t \geq \tau$}
        \STATE $\mathcal{T} \gets \mathcal{T} \cup \{(a_t, M_t)\}$
        \STATE $\text{success} \gets \text{True}$
        \STATE \textbf{break}
    \ELSE
        \STATE \text{Resample } $a_t$ \text{ by selecting alternative local maxima in } $\mathcal{D}(\cdot)$
    \ENDIF
\ENDFOR
\IF{\NOT $\text{success}$} \STATE \textbf{break} \COMMENT{Early stop if no significant improvement after $N$ retries} \ENDIF
\ENDFOR\RETURN $\mathcal{T}$
\end{algorithmic}
\end{algorithm}

The key innovation of our approach lies in an adaptive error-driven refinement mechanism combined with progress-constrained sampling. Rather than randomly sampling prompts, we leverage morphological analysis of prediction errors. We identify False Negative (FN) regions where the model under-segments and False Positive (FP) regions where it over-segments, then apply distance transforms to localize the most significant error clusters. This ensures that each corrective action addresses the most salient morphological defects. Furthermore, to guarantee trajectory quality, we enforce a constraint where each action must yield a measurable IoU improvement, filtering out ineffective interactions through an iterative retry mechanism. Our trajectory generation strategy differs primarily in the initial prompt $a_0$. In the Box-to-Point workflow, $a_0$ is a bounding box generated by extracting the axis-aligned rectangle of $M_{\text{target}}$ with a random jitter $\epsilon \sim \mathcal{U}(-\delta, \delta)$ to simulate human imprecision. In contrast, the Sequential-Click paradigm begins with a point prompt sampled from the centroid of the target mask:

\begin{equation}
a_0 = \frac{1}{|M_{\text{target}}|} \sum_{(x,y) \in M_{\text{target}}} (x, y) + \epsilon
\end{equation}

Regardless of initialization, all subsequent refinement steps at $t > 0$ follow a unified error-driven mechanism. The agent analyzes the prediction error $E_t = M_{\text{pred}}^{(t)} \oplus M_{\text{target}}$, decomposed into $\text{FN}_t = M_{\text{target}} \setminus M_{\text{pred}}^{(t)}$ and $\text{FP}_t = M_{\text{pred}}^{(t)} \setminus M_{\text{target}}$. By applying a distance transform $\mathcal{D}(\cdot)$, we identify the centroids of the largest error components. Corrective clicks are sampled as:
\begin{equation}
a_t = \begin{cases}\arg\max_{p \in \text{FN}_t} \mathcal{D}(p), & \text{if } |\text{FN}_t| > |\text{FP}_t| \\ 
\arg\max_{p \in \text{FP}_t} \mathcal{D}(p), & \text{otherwise}\end{cases}
\end{equation}
with positive labels for FN clicks and negative labels for FP clicks.To ensure the efficiency of synthesized trajectories, we implement a progress-constrained sampling mechanism. We require each simulated action to yield an incremental IoU gain ($\Delta \text{IoU}$) exceeding a predefined threshold $\tau$:

\begin{equation}
\Delta \text{IoU}_t = \text{IoU}(M_t, M_{\text{target}}) - \text{IoU}(M_{t-1}, M_{\text{target}}) \geq \tau
\end{equation}
where $\tau$ is typically set to 0.04. If a candidate action fails to satisfy this threshold, the simulator performs iterative resampling up to $N$ trials (default $N=5$) to identify a more constructive interaction. During each retry, we re-sample the click position within the error region by selecting alternative local maxima in the distance transform map or introducing controlled perturbations. If no valid action is found after $N$ trials, the trajectory generation is terminated early. This validation process ensures that the trajectory dataset $\mathcal{D}_{\text{traj}}$ is composed of high-quality, monotonically improving sequences. We set $K=5$ as the maximum number of refinement clicks, $\tau=0.04$ as the gain threshold, and $N=5$ as the maximum retry attempts. Bounding box jitter is sampled from $\mathcal{U}(-5, 5)$ pixels, while click jitter follows $\mathcal{N}(0, 2^2)$ pixels. The generated trajectories are stored in JSON format, recording each action $a_t$, intermediate mask $M_t$.  

Our implementation leverages MedSAM2 \cite{medsam2} and IMISNet \cite{IMISNet} as the base segmentation models $f_\theta$. Based on this trajectory generation pipeline, we initially generated 334,616 trajectories for both the Box-to-Point and Sequential-Click strategies using MedSAM2. To ensure the quality of the Supervised Fine-Tuning (SFT) data, we applied an IoU threshold of 0.7 to filter out low-performing samples. This resulted in 188,687 click-based and 260,446 box-based trajectories, totaling 449,133 high-quality samples. A similar procedure was applied using IMISNet, yielding 235,993 click trajectories and 283,522 box trajectories, for a total of 519,515 samples. These filtered trajectories served as the SFT data to train their respective models.

\subsection{Prompt Design}

The reasoning capabilities of MedSAM-Agent are governed by a structured system prompt that defines the agent's role, tool-use protocols, and decision-making heuristics. This prompt ensures that the MLLM interprets the medical image not as a static object, but as a dynamic environment requiring strategic intervention.

\subsubsection{Tool Specifications}
We provide the agent with a rigorous schema for tool calling. Specifically, the model is instructed to output its decisions in a standard JSON format, selecting from \texttt{add\_box}, \texttt{add\_point}, or \texttt{stop\_action}. To maintain procedural clarity, the system prompt enforces a ``Single Action per Turn" rule, requiring the model to observe the visual feedback from the segmentation backend before initiating the subsequent refinement step.

\subsubsection{System Prompt}
As for the system prompt, we first initialize the model as a professional segmentation annotator, grounding its objective in high-precision clinical mask creation. The prompt explicitly informs the agent that the mask is rendered as a ``semi-transparent green overlay", enabling the model to visually contrast the current prediction with the underlying anatomy to identify areas requiring refinement.

\begin{protocolbox}{Functional Tools (json)}
\begin{lstlisting}
[
  {
    "type": "function",
    "function": {
      "name": "add_bbox",
      "description": "Add a bounding box to initialize or refine the segmentation.",
      "parameters": {
        "type": "object",
        "properties": {
          "bbox_2d": { "type": "array", "items": {"type": "integer"}, "minItems": 4, "maxItems": 4, "description": "2D bounding box in [x1, y1, x2, y2] format" }
        },
        "required": ["bbox_2d"]
      }
    }
  },
  {
    "type": "function",
    "function": {
      "name": "add_point",
      "description": "Add a point to refine the mask (positive to include areas, negative to exclude areas).",
      "parameters": {
        "type": "object",
        "properties": {
          "point_2d": { "type": "array", "items": {"type": "integer"}, "minItems": 2, "maxItems": 2, "description": "2D coordinate point in [x, y] format" },
          "point_type": { "type": "string", "enum": ["positive", "negative"] }
        },
        "required": ["point_2d", "point_type"]
      }
    }
  },
  {
    "type": "function",
    "function": {
      "name": "stop_action",
      "description": "Stop the refinement process when the mask accurately covers the target object.",
      "parameters": { "type": "object", "properties": {} }
    }
  }
]
\end{lstlisting}
\end{protocolbox}

\begin{protocolbox}{System Prompt}
You are a professional segmentation annotator specializing in mask creation and refinement. Your core task is to segment the USER-SPECIFIED TARGET REGION from the provided image. No preliminary mask is available—you must first create an initial mask using the tool, then iteratively refine it to achieve pixel-level accuracy. The mask will be displayed as a semi-transparent green overlay; your goal is to ensure it exactly covers the entire target region and excludes all non-target areas (e.g., background, adjacent objects).

\vspace{1em}
\noindent \textbf{\# Tools}

\noindent You must call one function to assist with the user query. You are provided with function signatures within \texttt{<tools></tools>} XML tags:\\
\noindent \texttt{<tools>} \\
\noindent \textit{[The JSON-formatted tool signatures are inserted here]} \\
\texttt{</tools>}

\noindent For each function call, return a json object with function name and arguments within \texttt{<tool\_call></tool\_call>} XML tags:

\vspace{0.5em}
\begin{tcolorbox}[colback=white, boxrule=0.5pt, sharp corners, left=2mm, right=2mm, top=1mm, bottom=1mm]
\texttt{<tool\_call>} \\
\texttt{\{"name": <function-name>, "arguments": <args-json-object>\}} \\
\texttt{</tool\_call>}
\end{tcolorbox}

\vspace{0.5em}
\noindent Only use the provided functions to complete your task. Do not invent or assume any other functions. Carefully consider the current mask state before each action.
\end{protocolbox}

\subsubsection{Interaction Protocol}
The interaction protocol is structured as a closed-loop dialogue that transitions from initial localization to iterative refinement, mimicking a clinical workflow. The process begins with an initialization turn, where the agent receives the target specification and performs the first localization action. This is followed by subsequent refinement turns, during which the agent evaluates the visual feedback of the mask overlay against the anatomical structure, strategically placing positive or negative points to correct errors. The cycle continues until the agent, guided by its internal reward-optimized policy, determines that the segmentation has reached pixel-level accuracy and invokes the termination action to finalize the task efficiently.
\begin{protocolbox}{Initial Turn prompt}
\texttt{<image>} The target to be segmented is: \textbf{\{target\_description\}}. \\
Now, please analyze the original image, then decide your first action.
\end{protocolbox}

\begin{protocolbox}{Subsequent Turns prompt}
\texttt{<image>} Here is the updated mask after your previous action. Based on this, what is your next action? If the mask is now accurate, you can call \textbf{'stop\_action'} to finish.
\end{protocolbox}

Based on these prompts, we format the resulting interaction sequences into a standard Supervised Fine-Tuning (SFT) data structure to facilitate the model's cold-start training.

\subsection{Training Details}
The resolution of all datasets in our experiments is $1024 \times 1024$. During the construction of interaction trajectories, we map the spatial coordinates to absolute integers within a $[0, 1000]$ range to ensure compatibility with the Qwen3-VL \cite{Qwen2.5-VL} architecture. This normalization enables the model to effectively process spatial prompts and achieve high-precision interactive segmentation.

During the SFT stage, we utilize the Llama-Factory \cite{llamafactory} framework with a learning rate of $1\times10^{-5}$ and a total batch size of 64. The number of training epochs is 4. To maintain training efficiency and leverage pre-trained visual representations, we only update the parameters of the LLM backbone, while keeping the vision encoder and projector frozen. The training is accelerated using the DeepSpeed ZeRO-3 strategy.

The RL stage is implemented via the Verl framework \cite{verl}, using a learning rate of $1\times10^{-5}$, a batch size of 8, and a sampling size of 8 per prompt. The maximum interaction depth is set to 5 turns. To maximize training throughput, we employ an asynchronous multi-turn sampling mechanism integrated with the SGLang \cite{sglang} runtime, enabling efficient concurrent inference during the policy rollout phase. As for the hyper-parameters, we set $w_{iou}=w_{dice}=0.5$, $w_1=0.2$, $w_2=0.8$, $\lambda_1=0.1$, $\lambda_2=1.0$, and $\lambda_3=0.01$. All experiments were conducted on 8 $\times$ NVIDIA H20 (96G) GPUs. 

During the RL stage, given that medical segmentation requires a rigorous focus on fine-grained visual details, we deviate from the conventional reasoning-agent setup by omitting a dedicated \texttt{<think>} token. This design choice is driven by two primary objectives: first, to significantly accelerate inference speed by reducing token overhead; and second, to force the model to prioritize dense image-level features over textual deliberation. By bypassing the explicit thinking step, we encourage the agent to internalize spatial reasoning directly within its action space, ensuring that its decision-making is more tightly coupled with the immediate morphological characteristics of the lesion.
\section{Experiments Results}

\subsection{Comparison Experiment}
In our experiments, we compare our method with representative interactive segmentation frameworks, including the general-purpose SAM2~\cite{SAM2} and domain-specific models such as MedSAM2~\cite{medsam2} and IMISNet~\cite{IMISNet}. For these baselines, we report results using both \textit{Point} and \textit{Box} prompts. ``Point" results are generated using the center of the ground-truth mask, while ``Box" results utilize corresponding bounding boxes. The Box prompt performance is generally regarded as the empirical upper bound for single-turn interactive segmentation. Furthermore, we benchmark against SOTA MLLM-based methods, including four general-domain models (LISA~\cite{LISA}, GALLM~\cite{GALLM}, HyperSeg~\cite{hyperseg}, and Seg-R1~\cite{Seg-R1}) and three specialized medical MLLMs (MedPLIB~\cite{MedPLIB}, UniBioMed~\cite{UniBiomed}, and Citrus-V~\cite{citrusV}).

Table~\ref{tab:allres1} and Table~\ref{tab:allres2} demonstrate all results of 21 datasets across 6 medical modalities. The empirical results reveal that the performance of our autonomous agent is intrinsically linked to the foundational capabilities of the underlying segmentation model, yet it demonstrates a unique capacity to improve upon baseline results through strategic interaction. For instance, while specialized models like IMISNet demonstrate exceptional proficiency in CT datasets, their performance significantly decreases when applied to X-Ray, as observed in the Radiography \cite{chowdhury2020can} results. However, our multi-turn agentic framework addresses these limitations by iteratively identifying residual errors and applying corrective logic. By engaging in multiple rounds of interaction, the agent achieves performance that surpasses the inherent limits of a single-turn approach, thereby demonstrating the fundamental significance of multi-turn strategic engagement. This phenomenon suggests that MedSAM-Agent has internalized a sophisticated refinement logic that goes beyond simple tool invocation. By evaluating the segmentation state at each turn, the agent autonomously decides on corrective measures to eliminate hallucinations and under-segmentation. This active perception allows the agent to enhance raw model capabilities to reach the high precision required for clinical diagnostics, proving that autonomous, multi-turn strategic interaction is superior to passive, single-turn prompting paradigms.

\begin{table*}[htb]
\centering
\caption{\textbf{Quantitative comparison across CT, MRI, and Endoscopy.} Point and Box denote the interactive prompts derived from the ground-truth center points and bounding boxes, respectively, which are utilized to evaluate the single-round segmentation performance. \textbf{Bold} and \underline{underlined} values represent the best and second-best performance among non-interactive methods, respectively.}
\setlength{\tabcolsep}{2pt}
\begin{adjustbox}{width=\textwidth,center}
\begin{tabular}{llccccccccccc}
\toprule
 &  & \multicolumn{6}{c}{CT} & \multicolumn{3}{c}{MRI} & \multicolumn{2}{c}{Endoscopy} \\
\cmidrule(lr){3-8} \cmidrule(lr){9-11} \cmidrule(lr){12-13}
 &  & Flare22 & KiTS & LIDC-IDRI & BTCV & amos-CT & WORD & ACDC & LGG & amos-MRI & NeoPolyp & PolypGen \\
\midrule
\multirow{2}{*}{SAM2-Point} & Dice & 0.766 & 0.666 & 0.508 & 0.634 & 0.601 & 0.487 & 0.532 & 0.439 & 0.492 & 0.641 & 0.678 \\
 & IoU & 0.697 & 0.563 & 0.430 & 0.564 & 0.529 & 0.405 & 0.446 & 0.354 & 0.436 & 0.585 & 0.608 \\
\midrule
\multirow{2}{*}{SAM2-Box} & Dice & 0.920 & 0.818 & 0.756 & 0.890 & 0.869 & 0.772 & 0.809 & 0.865 & 0.839 & 0.934 & 0.911 \\
 & IoU & 0.863 & 0.720 & 0.648 & 0.821 & 0.796 & 0.674 & 0.718 & 0.775 & 0.752 & 0.887 & 0.856 \\
\midrule
\multirow{2}{*}{MedSAM-Point} & Dice & 0.831 & 0.655 & 0.763 & 0.795 & 0.751 & 0.686 & 0.556 & 0.869 & 0.577 & 0.844 & 0.800 \\
 & IoU & 0.746 & 0.528 & 0.645 & 0.700 & 0.660 & 0.558 & 0.498 & 0.779 & 0.476 & 0.762 & 0.711 \\
\midrule
\multirow{2}{*}{MedSAM-Box} & Dice & 0.913 & 0.836 & 0.773 & 0.890 & 0.888 & 0.781 & 0.847 & 0.893 & 0.821 & 0.933 & 0.916 \\
 & IoU & 0.847 & 0.735 & 0.668 & 0.815 & 0.814 & 0.674 & 0.766 & 0.813 & 0.725 & 0.884 & 0.859 \\
\midrule
\multirow{2}{*}{IMISNet-Point} & Dice & 0.845 & 0.677 & 0.268 & 0.819 & 0.745 & 0.692 & 0.753 & 0.622 & 0.660 & 0.798 & 0.764 \\
 & IoU & 0.766 & 0.554 & 0.221 & 0.725 & 0.638 & 0.577 & 0.634 & 0.485 & 0.548 & 0.704 & 0.659 \\
\midrule
\multirow{2}{*}{IMISNet-Box} & Dice & 0.911 & 0.826 & 0.524 & 0.870 & 0.869 & 0.829 & 0.857 & 0.860 & 0.830 & 0.924 & 0.885 \\
 & IoU & 0.852 & 0.740 & 0.443 & 0.797 & 0.794 & 0.736 & 0.767 & 0.766 & 0.741 & 0.871 & 0.821 \\
\midrule
\multirow{2}{*}{LISA} & Dice & 0.116 & 0.147 & 0.006 & 0.062 & 0.097 & 0.098 & 0.069 & 0.185 & 0.076 & 0.271 & 0.293 \\
 & IoU & 0.070 & 0.086 & 0.003 & 0.036 & 0.057 & 0.055 & 0.037 & 0.107 & 0.044 & 0.220 & 0.225 \\
\midrule
\multirow{2}{*}{GLAMM} & Dice & 0.128 & 0.110 & 0.005 & 0.078 & 0.106 & 0.103 & 0.050 & 0.188 & 0.078 & 0.204 & 0.285 \\
 & IoU & 0.079 & 0.061 & 0.003 & 0.048 & 0.066 & 0.059 & 0.027 & 0.109 & 0.046 & 0.149 & 0.208 \\
\midrule
\multirow{2}{*}{HyperSeg} & Dice & 0.117 & 0.175 & 0.006 & 0.068 & 0.099 & 0.121 & 0.160 & 0.183 & 0.117 & 0.392 & 0.400 \\
 & IoU & 0.089 & 0.129 & 0.003 & 0.046 & 0.067 & 0.078 & 0.102 & 0.119 & 0.083 & 0.344 & 0.341 \\
\midrule
\multirow{2}{*}{Seg-R1} & Dice & 0.113 & 0.159 & 0.009 & 0.087 & 0.102 & 0.124 & 0.134 & 0.168 & 0.076 & 0.515 & 0.551 \\
 & IoU & 0.080 & 0.104 & 0.005 & 0.055 & 0.064 & 0.076 & 0.081 & 0.098 & 0.045 & 0.457 & 0.485 \\
\midrule
\multirow{2}{*}{MedPLIB} & Dice & 0.046 & 0.033 & 0.001 & 0.083 & 0.054 & 0.096 & 0.118 & 0.318 & 0.096 & 0.123 & 0.142 \\
 & IoU & 0.035 & 0.021 & 0.001 & 0.058 & 0.038 & 0.075 & 0.088 & 0.252 & 0.070 & 0.092 & 0.102 \\
\midrule
\multirow{2}{*}{UniBiomed} & Dice & 0.825 & 0.755 & 0.502 & 0.735 & 0.822 & 0.707 & 0.827 & \textbf{0.863} & \textbf{0.732} & \textbf{0.851} & 0.705 \\
 & IoU & 0.747 & 0.670 & 0.405 & 0.646 & 0.738 & 0.599 & 0.738 & \textbf{0.776} & \textbf{0.635} & \textbf{0.784} & 0.623 \\
\midrule
\multirow{2}{*}{Citrus-V} & Dice & 0.480 & 0.470 & 0.221 & 0.428 & 0.262 & 0.314 & 0.438 & 0.305 & 0.235 & 0.549 & 0.783 \\
 & IoU & 0.417 & 0.377 & 0.170 & 0.370 & 0.220 & 0.258 & 0.326 & 0.254 & 0.194 & 0.510 & 0.722 \\
\midrule
\multirow{2}{*}{Ours-IMISNet} & Dice & \textbf{0.848} & \textbf{0.791} & 0.406 & \textbf{0.780} & \textbf{0.828} & \textbf{0.741} & \textbf{0.836} & 0.828 & 0.685 & 0.813 & 0.793 \\
 & IoU & \textbf{0.779} & \textbf{0.710} & 0.340 & \textbf{0.702} & \textbf{0.751} & \textbf{0.642} & \textbf{0.743} & 0.728 & 0.583 & 0.753 & 0.719 \\
\midrule
\multirow{2}{*}{Ours-MedSAM2} & Dice & 0.836 & 0.703 & \textbf{0.523} & 0.773 & 0.805 & 0.681 & 0.816 & 0.844 & 0.719 & 0.814 & \textbf{0.809} \\
 & IoU & 0.761 & 0.597 & \textbf{0.448} & 0.690 & 0.723 & 0.579 & 0.726 & 0.755 & 0.623 & 0.753 & \textbf{0.735} \\
\bottomrule
\end{tabular}
\end{adjustbox}
\label{tab:allres1}
\end{table*}

\begin{table*}[htb]
\centering
\caption{\textbf{Quantitative comparison across X-Ray, Ultrasound, and Fundus.} Point and Box denote the interactive prompts derived from the ground-truth center points and bounding boxes, respectively, which are utilized to evaluate the single-round segmentation performance. \textbf{Bold} and \underline{underlined} values represent the best and second-best performance among non-interactive methods, respectively.}
\setlength{\tabcolsep}{1.5pt}
\begin{adjustbox}{width=\textwidth,center}
\begin{tabular}{llcccccccccc}
\toprule
 &  & \multicolumn{6}{c}{X-Ray} & \multicolumn{3}{c}{Ultrasound} & \multicolumn{1}{c}{Fundus} \\
\cmidrule(lr){3-8} \cmidrule(lr){9-11} \cmidrule(lr){12-12}
 &  & CXRMask & Rad-LO & Rad-N & Rad-VP & Rad-COVID & CDD-CESM & BreastUS & LiverUS & FH-PS-AOP & REFUGE \\
\midrule
\multirow{2}{*}{SAM2-Point} & Dice & 0.647 & 0.546 & 0.627 & 0.536 & 0.582 & 0.339 & 0.685 & 0.557 & 0.341 & 0.362 \\
 & IoU & 0.533 & 0.409 & 0.499 & 0.392 & 0.454 & 0.246 & 0.582 & 0.424 & 0.217 & 0.265 \\
\midrule
\multirow{2}{*}{SAM2-Box} & Dice & 0.940 & 0.790 & 0.928 & 0.783 & 0.890 & 0.628 & 0.889 & 0.849 & 0.876 & 0.717 \\
 & IoU & 0.890 & 0.667 & 0.869 & 0.667 & 0.809 & 0.497 & 0.811 & 0.745 & 0.782 & 0.574 \\
\midrule
\multirow{2}{*}{MedSAM-Point} & Dice & 0.887 & 0.878 & 0.956 & 0.922 & 0.941 & 0.477 & 0.808 & 0.421 & 0.543 & 0.414 \\
 & IoU & 0.828 & 0.796 & 0.919 & 0.861 & 0.894 & 0.360 & 0.711 & 0.286 & 0.405 & 0.346 \\
\midrule
\multirow{2}{*}{MedSAM-Box} & Dice & 0.946 & 0.912 & 0.976 & 0.903 & 0.955 & 0.692 & 0.918 & 0.856 & 0.896 & 0.846 \\
 & IoU & 0.899 & 0.842 & 0.953 & 0.832 & 0.918 & 0.572 & 0.851 & 0.752 & 0.815 & 0.743 \\
\midrule
\multirow{2}{*}{IMISNet-Point} & Dice & 0.719 & 0.444 & 0.764 & 0.462 & 0.694 & 0.386 & 0.726 & 0.257 & 0.483 & 0.677 \\
 & IoU & 0.590 & 0.304 & 0.647 & 0.316 & 0.564 & 0.271 & 0.607 & 0.171 & 0.335 & 0.528 \\
\midrule
\multirow{2}{*}{IMISNet-Box} & Dice & 0.948 & 0.639 & 0.916 & 0.465 & 0.838 & 0.430 & 0.873 & 0.209 & 0.625 & 0.841 \\
 & IoU & 0.901 & 0.535 & 0.859 & 0.368 & 0.757 & 0.349 & 0.791 & 0.148 & 0.509 & 0.733 \\
\midrule
\multirow{2}{*}{LISA} & Dice & 0.243 & 0.621 & 0.220 & 0.487 & 0.187 & 0.177 & 0.218 & 0.615 & 0.321 & 0.036 \\
 & IoU & 0.184 & 0.469 & 0.163 & 0.325 & 0.126 & 0.111 & 0.147 & 0.466 & 0.227 & 0.019 \\
\midrule
\multirow{2}{*}{GLAMM} & Dice & 0.192 & 0.504 & 0.178 & 0.473 & 0.173 & 0.190 & 0.315 & 0.611 & 0.282 & 0.027 \\
 & IoU & 0.136 & 0.344 & 0.121 & 0.325 & 0.117 & 0.121 & 0.231 & 0.463 & 0.189 & 0.014 \\
\midrule
\multirow{2}{*}{HyperSeg} & Dice & 0.329 & 0.413 & 0.330 & 0.463 & 0.323 & 0.157 & 0.536 & 0.609 & 0.157 & 0.023 \\
 & IoU & 0.239 & 0.266 & 0.246 & 0.305 & 0.227 & 0.103 & 0.450 & 0.458 & 0.100 & 0.012 \\
\midrule
\multirow{2}{*}{Seg-R1} & Dice & 0.446 & 0.583 & 0.383 & 0.522 & 0.385 & 0.186 & 0.625 & 0.613 & 0.233 & 0.455 \\
 & IoU & 0.342 & 0.445 & 0.312 & 0.372 & 0.313 & 0.121 & 0.521 & 0.463 & 0.162 & 0.325 \\
\midrule
\multirow{2}{*}{MedPlib} & Dice & 0.225 & 0.086 & 0.076 & 0.030 & 0.156 & 0.116 & 0.135 & 0.088 & 0.042 & 0.297 \\
 & IoU & 0.154 & 0.049 & 0.048 & 0.017 & 0.106 & 0.073 & 0.098 & 0.048 & 0.023 & 0.201 \\
\midrule
\multirow{2}{*}{UniBiomed} & Dice & 0.936 & 0.887 & 0.959 & 0.871 & 0.894 & 0.357 & \textbf{0.828} & 0.620 & 0.761 & 0.794 \\
 & IoU & 0.885 & 0.804 & 0.923 & 0.782 & 0.830 & 0.262 & \textbf{0.743} & 0.470 & 0.640 & 0.668 \\
\midrule
\multirow{2}{*}{Citrus-V} & Dice & 0.694 & 0.702 & 0.713 & 0.799 & 0.653 & 0.186 & 0.324 & 0.040 & 0.038 & 0.782 \\
 & IoU & 0.601 & 0.600 & 0.623 & 0.716 & 0.542 & 0.136 & 0.237 & 0.021 & 0.022 & 0.655 \\
\midrule
\multirow{2}{*}{Ours-IMISNet} & Dice & \textbf{0.947} & 0.832 & 0.941 & 0.843 & 0.902 & 0.324 & 0.788 & 0.432 & 0.756 & 0.803 \\
 & IoU & \textbf{0.899} & 0.729 & 0.892 & 0.747 & 0.829 & 0.242 & 0.699 & 0.294 & 0.641 & 0.678 \\
\midrule
\multirow{2}{*}{Ours-MedSAM2} & Dice & 0.930 & \textbf{0.900} & \textbf{0.973} & \textbf{0.874} & \textbf{0.948} & \textbf{0.375} & 0.805 & \textbf{0.777} & \textbf{0.798} & \textbf{0.813} \\
 & IoU & 0.876 & \textbf{0.828} & \textbf{0.950} & \textbf{0.796} & \textbf{0.906} & \textbf{0.291} & 0.719 & \textbf{0.647} & \textbf{0.689} & \textbf{0.692} \\
\bottomrule
\end{tabular}
\end{adjustbox}
\label{tab:allres2}
\end{table*}

\subsection{Case Study}
\begin{figure}[t]
  \centering 
  \includegraphics[width=0.9\linewidth]{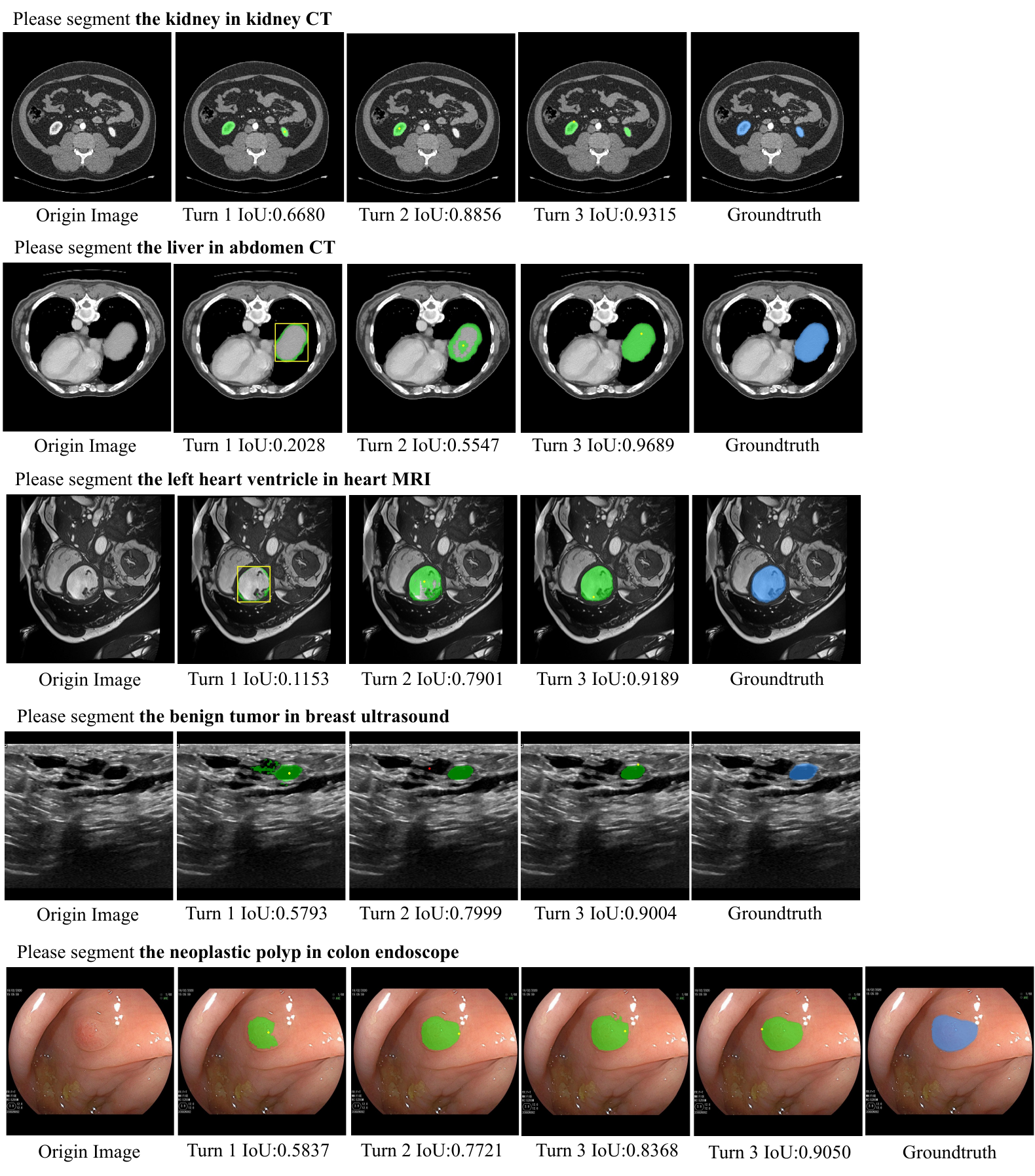}
   \caption{\textbf{Case study.} Yellow boxes indicate bounding box prompts, yellow points represent positive clicks, and red points denote negative clicks.}
   \label{fig:case_study}
\end{figure}

In this section, we provide qualitative examples of MedSAM-Agent across various medical imaging modalities to demonstrate its interactive reasoning capabilities (Fig.~\ref{fig:case_study}). Unlike traditional single-turn models, our agent engages in a multi-step refinement process, iteratively evaluating the current segmentation mask and adjusting its strategy based on visual feedback. In these visualizations, yellow prompts indicate positive clicks or bounding boxes used to define the target anatomy, while red prompts represent negative clicks deployed to exclude false-positive regions or refine over-segmented boundaries.

As shown in the sequential turns, MedSAM-Agent typically initiates the process with a global constraint and subsequently applies local corrective measures to eliminate hallucinations and fill in under-segmented areas. This autonomous refinement loop effectively mimics a clinician's iterative workflow, proving that the model can successfully internalize sophisticated decision heuristics to achieve superior mask fidelity in challenging clinical environments.

\section{Future Works}
In future research, we plan to extend the MedSAM-Agent framework along the following three strategic dimensions:

\textbf{Expansion to Volumetric Modalities.} While the current study focuses on 2D image slices, many clinical diagnostic tasks rely on 3D volumetric data, such as CT and MRI scans. We aim to extend the agent's action space and reinforcement learning environment to handle 3D spatial contexts. This transition will enable the agent to capture cross-slice anatomical continuity, providing more consistent and clinically accurate volumetric segmentations.

\textbf{Development of a Unified multi-modal Agent.} While our work focuses on the specialized task of interactive segmentation, the proposed autonomous decision-making framework is inherently extensible. In the future, we intend to evolve MedSAM-Agent into a unified medical AI assistant capable of performing diverse medical imaging tasks (e.g., Medical VQA, lesion classification, and automated report generation) within a single, cohesive architecture. By treating these distinct tasks as specialized tool-use actions, we can integrate new capabilities without altering the core agentic framework. This evolution aims to provide a comprehensive assistant that supports the entire clinical workflow, seamlessly transitioning between perception, reasoning, and high-precision interaction.

\textbf{Enhancement of Computational Efficiency.} A potential limitation of multi-turn iterative paradigms is the increased cumulative latency during inference. In future work, we plan to explore more efficient architectures and sampling strategies to mitigate this overhead. Specifically, we will investigate the integration of early-exit mechanisms to terminate redundant computation when high-confidence masks are achieved and leverage speculative decoding or KV-cache optimization techniques tailored for multi-modal agents. These advancements will ensure that the superior precision of MedSAM-Agent is delivered with the near-real-time responsiveness required for high-throughput clinical applications.
\end{document}